  \documentclass[letterpaper, 10pt, twocolumn]{article}
  \usepackage{clean}
  \shorttitle{MPIC with GPs for Interaction}

\usepackage{graphicx}
\usepackage{subfig}
\graphicspath{{figs/}}

\usepackage{amsmath,amssymb,amsfonts}
\usepackage{algorithm2e}
\usepackage{xcolor}
\usepackage[colorlinks=true, linkcolor=black, urlcolor=cyan, filecolor=black, citecolor=black]{hyperref}
\usepackage{url}
\usepackage{amsmath,bm}
\usepackage{multicol, multirow, makecell} 
\usepackage{comment}

\usepackage{cite} 

\title{Model Predictive Impedance Control with Gaussian Processes for Human and Environment Interaction}

\author{Kevin Haninger,~Christian Hegeler,~and~Luka Peternel
\thanks{K. Haninger and C. Hegeler are with the Department of Automation at Fraunhofer IPK, Berlin, Germany. L. Peternel is with the Cognitive Robotics Department, Delft Univerity of Technology, Delft, The Netherlands. Corresponding email: {\tt kevin.haninger@ipk.fraunhofer.de}}
\thanks{This project has received funding from the European Union's Horizon 2020 research and innovation programme under grant agreement Nos  820689 — SHERLOCK and 101058521 - CONVERGING.}}%

\begin{document}

\maketitle

\begin{abstract}
Robotic tasks which involve uncertainty--due to variation in goal, environment configuration, or confidence in task model--may require human input to instruct or adapt the robot. In tasks with physical contact, several existing methods for adapting robot trajectory or impedance according to individual uncertainties have been proposed, e.g., realizing intention detection or uncertainty-aware learning from demonstration. However, isolated methods cannot address the wide range of uncertainties jointly present in many tasks. 

To improve generality, this paper proposes a model predictive control (MPC) framework which plans both trajectory and impedance online, can consider discrete and continuous uncertainties, includes safety constraints, and can be efficiently applied to a new task. This framework can consider uncertainty from: contact constraint variation, uncertainty in human goals, or task disturbances. An uncertainty-aware task model is learned from a few ($\leq3$) demonstrations using Gaussian Processes. This task model is used in a nonlinear MPC problem to optimize robot trajectory and impedance according to belief in discrete human goals, human kinematics, safety constraints, contact stability, and frequency-domain disturbance rejection. This MPC formulation is introduced, analyzed with respect to convexity, and validated in co-manipulation with multiple goals, a collaborative polishing task, and a collaborative assembly task.   
\end{abstract}

\section{Introduction}
Manipulation with physical contact is essential for robotics, where impedance control\footnote{Impedance is used to refer to general physical interaction control, i.e., both impedance and admittance control.} is often paramount to robust and safe execution. This is especially critical in tasks that involve physical human-robot interaction (HRI), where the robot compliance allows the human co-worker to safely affect robot motion for teaching or online collaboration. Impedance control defines the characteristics of physical interaction by stiffness, damping, and inertia parameters \cite{hogan1985}. These parameters must be appropriately adapted to an application, with a large number of proposed methods to determine appropriate impedance parameters. Nevertheless, in practice, the tuning of these parameters is often done with ad-hoc or application-specific methods.

Some recent works propose generalizable approaches to adapting impedance parameters based on machine learning and human-in-the-loop approaches. A common theme in impedance adaptation is the degree of uncertainty. From the literature, we identify five key categories related to uncertainty and impedance control. 1) In learning from demonstration (LfD), uncertainty derived from the \textit{variability in demonstrations} of trajectories can be used to infer lower robot stiffness, since the human demonstrator does not seem to care about precision in that section of motion \cite{calinon2010,kronander2014}. 2) In LfD, the robot's available \textit{confidence in the skill} is used, where stiffness can be reduced when entering unexplored regions as actions might be unsafe \cite{franzese2021}. 3) When faced with uncertainty in \textit{environment constraints} in task execution, the robot should have lower stiffness in order not to produce dangerously high forces when the contact is unexpected \cite{walker2011,ajoudani2012,peternel2018a}. 4) In case of uncertain \textit{external perturbations} the robot should increase the impedance to ensure the accuracy of position tracking task \cite{yang2011a,naceri2021learning}. 5) When there is uncertainty in \textit{human goal} during collaborative tasks, the robot should be compliant to let them determine a specific degree of freedom (DoF). For example, when the human should lead the collaborative carrying of an object \cite{duchaine2007}, or when the human should guide a polishing machine that is carried by the robot \cite{peternel2018}.

Nevertheless, scaling up impedance control to different tasks, conditions and environments while accounting for human goals in real-time is still a major challenge in physical HRI. The existing methods typically focus on either one or a couple of these uncertainty categories, although many tasks feature several of them. We aim to develop a framework which can account for multiple types of uncertainties for a unified physical HRI approach. The goal is for robots to be able to plan trajectories and adapt impedance for semi-structured tasks in physical contact with humans and environments, without needing an analytical model of environment/human or large amounts of data. To attain that goal, we propose a planning and optimization framework with probabilistic models of external forces over robot state. This acts as an outer loop controller, adapting robot trajectory and impedance parameters. The main contributions of this work are:
\begin{itemize}
    \item A taxonomy of uncertainty in physical interaction, including their sources, consequences for safety and performance, and a unified problem formulation to address these with adaptive impedance control.
    \item Unlike existing approaches that use GPs as direct robot reference behaviour generators (e.g., \cite{franzese2021ilosa}), we use them as intermediate task models which combine with the robot impedance in a Model Predictive Control (MPC) problem, allowing explicit objectives and constraints which depend on the trajectory uncertainty.
    \item Unlike existing sampling-based approaches to adapting impedance online (e.g., \cite{roveda2020}), we propose the first gradient-based optimization of impedance, which allows constraints such as maximum contact force and contact stability.
    \item The approach is experimentally validated, allowing human goal inference, increasing impedance damping for contact stability, and reducing impedance in directions of task variation, with $10-20$ Hz MPC rates. The code\footnote{\url{https://gitlab.cc-asp.fraunhofer.de/hanikevi/gp-mpc-impedance-control}} and video\footnote{\url{https://youtu.be/Of2O3mHfM94}} are available.
\end{itemize} 

The uncertainty category 1) related to variability in demonstrations can be accounted for in the proposed framework through GP covariance, but we note this can produce unsafe impedance skills when the low demonstration variance arises from contact with the environment \cite{peternel2018a}. The proposed framework also addresses the uncertainty category 2) related to confidence in skill by leveraging on GP model uncertainty in cost function and optimization constraints. The uncertainty category 3) related to environment constraints is handled by the increase in GP covariance induced by fitting hyperparameters on data with contact variance. An additional frequency-domain perturbation model is used to resolve the uncertainty category 4) related to external perturbations. The uncertainty category 5) related to human goal estimation is addressed by taking the expectation over the belief in human goal, inferred from the GP.

\begin{figure}[t]
    \centering
    \includegraphics[width=0.45\textwidth]{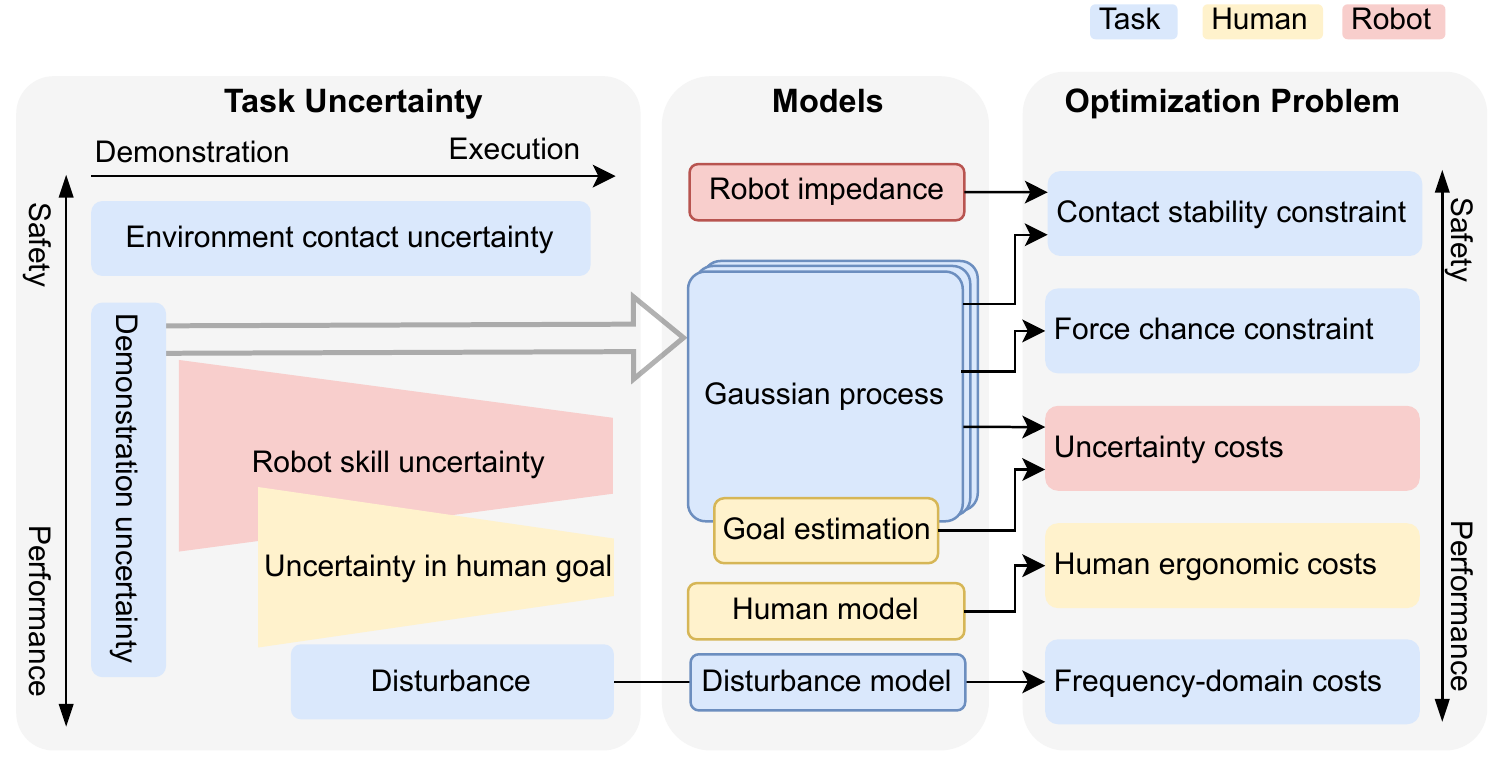} 
    \caption{Sources of uncertainty during both demonstration and execution (left), how they are modelled (center), and how they enter the MPC framework (right)}
    \label{fig:uncertainty}
\end{figure}

A preliminary study was presented at the 2022 IEEE International Conference on Robotics and Automation \cite{haninger2021a}. This paper significantly expands upon the previous study by an extended theoretical formulation of the method that accounts for all identified uncertainties, adding two types of MPC constraints, a new theoretical analysis of impedance optimization, and a new experimental evaluation with three additional practical tasks involving physical HRI (collaborative rail assembly, double peg-in-the-hole, and polishing) where we examine new aspects (task uncertainty, goal uncertainty, and human ergonomics).

The remainder of the paper is structured as follows (see also Fig. \ref{fig:overview} for overview). The related work is given in Section \ref{sec:related_work}. The modeling using GP is described in Section \ref{sec:models}. The MPC problem formulation is provided in Section \ref{sec:mpc}. This is followed by a theoretical analysis in Section \ref{sec:analysis}. Finally, experimental validation on practical tasks is described in Section \ref{sec:validation}.

\section{Related work \label{sec:related_work}}
Here we examine the proposed method with respect to related work in the literature. We first outline the work done in physical human-robot interaction where human intention modeling plays a critical role in seamless collaboration. Then we proceed to robot impedance adaptation, which is a fundamental part of the robot's low-level actions as a result of the inferred intention. Finally, we review planning in contact tasks.

\subsection{Physical human-robot interaction}
An important challenge in physical HRI is the ability of robots to infer human co-worker's goals and states. To address this challenge, several recent works have proposed inference techniques and applications. The study in \cite{takagi2017} showed that robot adaptation to the human task execution can improve task-related performance. Robot adaption to the human state can also improve ergonomics \cite{peternel2018}. Finally, user satisfaction can also be improved by robot adaptation \cite{corteville2007}. Nevertheless, to adapt to these different aspects, the robot should be able to infer the human state, which often requires complex sensory setups \cite{ajoudani2017,maurice2019}. However, complex measurement systems are not always available or feasible for practical applications, thus some specific states can be modeled, such as the desired motion of the robot/payload \cite{demiris2007, li2014, wang2017, kang2019}, human actions \cite{jain2018}, preferences \cite{jain2015,bajcsy2017}, human physical fatigue \cite{peternel2018}, cognitive stress \cite{messeri2021}, and suitability of shared workspace \cite{vahrenkamp2016,gopinathan2017,mansfeld2018,peternel2021}.

For seamless collaboration, the adaption of the robot to the human should be prompt, and thus the human intent or goal must be inferred during the task execution \cite{benamor2014,peternel2018}. Human intent is often considered as a continuous variable \cite{takagi2017,kang2019}, however, it can also reflect discrete changes in the task \cite{khoramshahi2019}, e.g., a collection of possible goals. Discrete human state has been considered for virtual fixtures \cite{raiola2018}, Dynamic Movement Primitives (DMPs) \cite{khoramshahi2019}, and impedance control \cite{rozo2016}. 

When the robot has the knowledge of the human state and intent, it must respond promptly with appropriate actions to facilitate the collaborative task execution. This typically means generation of motion trajectories and often simultaneous impedance adaptation. The trajectories can be chosen to optimize task-related objectives: reducing trajectory jerk \cite{maeda2001, dimeas2015}, minimize positioning error \cite{yang2011a,gribovskaya2011,lee2012}, or render appropriate velocity response \cite{tsumugiwa2002,duchaine2012}. In addition, human operator-related objectives can be optimized as well, such as minimizing interaction forces \cite{lamy2009} or metabolic cost \cite{koller2016}. A common rule is that the robot can take over aspects of the task that require precision, while the human can handle adaption to variations. The `minimum intervention principle' \cite{medina2012} follows this rule, where uncertain DOF should have a lower stiffness \cite{calinon2014, pignat2017}, which can also be interpreted as a risk-sensitive control. Nevertheless, such an approach is not reasonable for all interactive tasks that involve physical constraints \cite{peternel2018a}.

The literature can address the human-related properties (e.g., actions, preferences, ergonomics, etc.), real-time intention processing, and uncertainty in an isolated manner. However, a unified framework that can effectively combine all these aspects for practical manufacturing tasks is still missing.

\begin{table*}[h]
\renewcommand{\arraystretch}{1.6} 
\begin{center}
\caption{Selected prior work in adaptation of impedance, organized according to motivating principle. Robotic applications to co-manipulpation and contact are shown, as well as the targeted degree of robot autonomy.}
\begin{tabular}{r l  l  c c c} 
 \multicolumn{2}{c}{Principle} & Robot impedance should: & Co-manip. & Env. Contact & Autonomy \\ [0.5ex] 
 \hline
 \multirow{3}{*}{\rotatebox[origin=c]{90}{Dynamics}} & complementarity \cite{hogan1988} & be inverse of environment & - & -  & - \\ 
 \cline{2-6}
  & stability \cite{burdet2001,naceri2021learning} & stabilize unstable dyn & \cite{yang2011a} & \cite{li2018}  & full \\
  \cline{2-6}
  & well-damped \cite{tsumugiwa2002} & reduce oscillation or jerk & \cite{tsumugiwa2002, duchaine2007} & \cite{ferraguti2019} & interact, full \\
 \hline
 \multirow{3}{*}{\rotatebox[origin=c]{90}{\makecell{Optimal \\ Control}}} & min intervention \cite{calinon2010} & stiff where low variance & \cite{kormushev2010, medina2012, calinon2014a} &  \makecell{\cite{calinon2014a}\\ safety$^{a}$ \cite{peternel2018a}}  & interact, full\\
 \cline{2-6}
 & game theory \cite{li2019} & min coupled cost function & \cite{li2019b} & - & interact \\
 \cline{2-6}
  & reinf. learning & min cost function & \cite{dimeas2015} & \cite{stulp2012, chang2022} & full \\
 \hline
 \multirow{3}{*}{\rotatebox[origin=c]{90}{\makecell{Transfer\\ Hum Imp}}} & human hand imp \cite{kronander2014} & match human impedance & - & \cite{tang2015} & full \\
 \cline{2-6}
 & co-contraction \cite{peternel2017} & prop or inv to co-contraction & - & \cite{peternel2017} & interact \\
  \cline{2-6}
 & demo dynamics \cite{rozo2016} & which explains demos & \cite{rozo2016}  & - & interact \\
 \hline
\end{tabular}
\label{tab:imp-adapt}
\end{center}
\footnotesize{$^a$This relates to the robot being stiff while being rigidly constrained by a potentially high-stiffness environment, which can cause contact instability.}
\end{table*}

\subsection{Impedance adaptation}
Physical HRI requires a safe and robust low-level control approach to account for the uncertainty of practical tasks \cite{peternel2017}. Classic position or force controllers often fail to perform well in such complex settings since they only focus on controlling either one of the variables, but not the characteristics of the interaction that define their relationship, i.e., impedance. Impedance adaption is empirically important for safety and robustness in manipulation, and has been shown to improve sample efficiency of higher-level learning control \cite{ulmer2021}. Accordingly, a wide range of techniques has been proposed for adjusting impedance parameters. Recent surveys provide a thorough overview of this field \cite{abu-dakka2020, sharifi2021}, so we present only an abbreviated summary in Table \ref{tab:imp-adapt}.

Dynamics-based objectives are often based on continuous-time dynamic models, considering factors like stability and damping of the coupled system. A challenge is that environment dynamics (e.g., stiffness) are application-specific. In some cases, the environment dynamics are known or controlled \cite{burdet2001}, in other cases properties of the dynamics (e.g., damping) can be found from trajectories \cite{ferraguti2015}.

Other approaches plan to minimize a general cost function, which may include the human or environment state \cite{li2019b}. Such approaches allow consideration of uncertainty, but typically also require models of humans or the environment, which are typically simplified such as assuming human force is a PD controller \cite{medina2012}. Data-driven methods such as reinforcement learning can be applied \cite{stulp2012, chang2022}, but often require lots of trial and error. 

An approach in \cite{franzese2021} uses GP models to encode demonstrated desired robot movements and then leverages on the measure of uncertainty in the modeled skill to modulate the stiffness through a direct expression. If the robot experiences unexpected external perturbations, the impedance controller forces the motion into the regions of higher certainty to be more confident in the given skill. If the robot still ends up in a region of high uncertainty, the stiffness is reduced as there might be insufficient confidence in the skill to safely perform the movements there. Nevertheless, there was no higher-level planner (e.g., MPC) to exploit given learned trajectories to generate new ones.

From this, we conclude that there is a wide range of objectives and models--it seems unlikely that a single principle would be suited for all or a large majority of these areas. Accordingly, we propose a framework for optimizing impedance, capable of accommodating a range of models and objectives.

\subsection{Planning in contact tasks}
Intention modeling and low-level impedance control alone can work well for simplified tasks. However, for a more flexible robot behavior, a higher-level planner is required. MPC is becoming increasingly popular for manipulation \cite{rubagotti2019,roveda2020}, and is already a standard tool for locomotion and whole-body control, capable of planning through contact models with either linear complementarity \cite{manchester2020a} or switching dynamic modes \cite{sleiman2021}. Contact can be considered as fully rigid constraints, using time-stepping approaches to resolve the contact impulse over an integrator time step \cite{stewart2000}, avoiding the issue of unbounded contact forces in rigid contact. A dynamic model for the environment has also been proposed for manipulation \cite{minniti2021}, relaxing kinematic constraints into a stiffness. In most contact planning situations, signed distance functions are needed--typically, this requires geometry information typically given by CAD, although it has been learned on simplified objects \cite{pfrommer2020}. 

We make two observations from the perspective of manipulation: 1) signed distance functions are difficult with the complex geometries typical in assembly and manipulation tasks, and 2) compliance is both common and critical--be it physical or by control. To address this, we use a compliant contact model here (regressing contact force over robot position) which does not require a signed distance function, and we consider the optimization of impedance parameters in the MPC problem.  

\begin{figure}[t]
    \centering
    \includegraphics[width=0.45\textwidth]{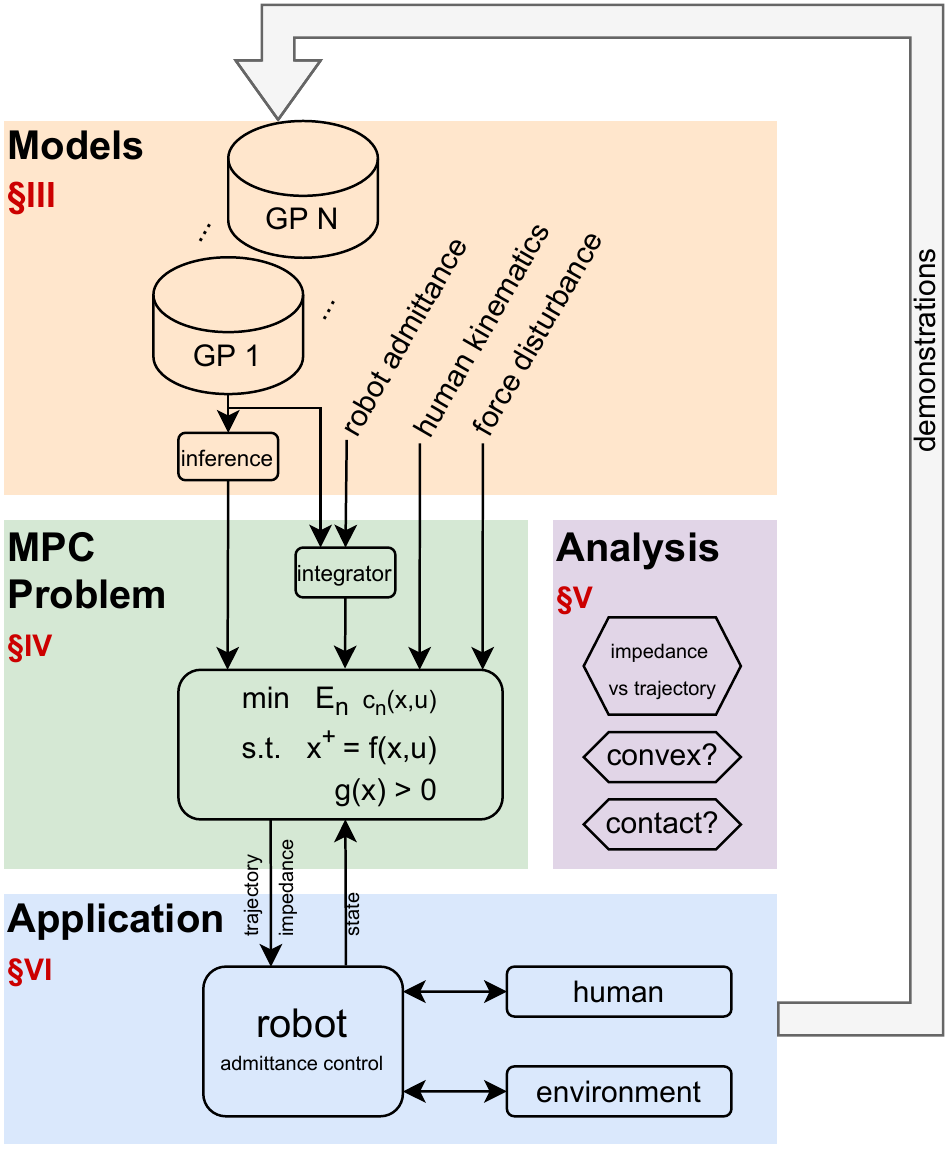}
    \caption{Overview of the proposed approach, with corresponding paper sections in red. Initially, human demonstrations are used to model human and task forces with Gaussian Processes. These models are then used in a model predictive controller that accounts for robot impedance and trajectories, and includes costs and constraints related to the task and human. The MPC controller runs online, updating trajectory and impedance parameters for a lower-level impedance controller.}
    \label{fig:overview}
\end{figure}

\section{Models\label{sec:models}}
This section presents the models used: GP models of force, human kinematics, robot admittance, and force disturbance. This aspect is highlighted by the orange part in Fig. \ref{fig:overview}.

\subsection{Force models}
For a robot with a tool-center point (TCP) pose $\bm{x}$, we consider three types of external forces:

\noindent\textbf{Human forces} $\bm{f}^h(\bm{x},n)$ are a function of robot state and human mode $n$, where $n\in[1\dots N]$ indexes possible human behavior (e.g., goals). 

\noindent\textbf{Contact forces} $\bm{f}^c(\bm{x},\bm{x}^e)$ depend on an environment configuration $\bm{x}^e$ which is typically unmeasured and may vary between iterations, such as a contact or constraint position. 

\noindent\textbf{Disturbance forces} $\bm{f}^d(t)$ such as polishing, cutting, or other process forces which should not affect robot position. We assume they can be represented in a frequency-domain model $\bm{F}^d(j\omega)$, e.g., they are predominantly high-frequency.

We use one force/torque (F/T) sensor, such that demonstrations measure total force $\bm{f} = \bm{f}^h+\bm{f}^c+\bm{f}^d$. While multiple F/T sensors can separate human and environment forces \cite{tang2015, chang2022}, a single F/T sensor reduces complexity. A disadvantage of a single F/T sensor is that in co-manipulation with contact, contact may be more difficult to identify, although this can be partly addressed by how the data is collected. Another disadvantage of this approach is that forces are regressed over position, which may be challenging in stiff contact. We validate both these aspects in Section \ref{sec:analysis_contact} and Fig. \ref{fig:contact_modeling}.  

\subsubsection{GP Models}
GPs are used to regress force over pose for each mode $n$, based on a dataset of measured pose and force $\mathcal{D} = \left\{\left\{\bm{x}_1,\bm{f}_1\right\}\dots,\left\{\bm{x}_M,\bm{f}_M\right\}\right\}$. We fit a GP Model \cite[p13]{rasmussen2006} that gives a Gaussian variable $\hat{\bm{f}}(\bm{x})\sim\mathcal{N}(\bm{\mu}_f(\bm{x}), \bm{\Sigma}_f(\bm{x}))$ for a pose $\bm{x}$, given a mean function $\bm{m}(\bm{x}) \in \mathbb{R}^6$, kernel function $k(\bm{x},\bm{x}')\in \mathbb{R}$, and observed data $\mathcal{D}$. The GP models for each dimension of $\bm{f}$ are fit independently and the covariance $\bm{\Sigma}_f$ is therefore diagonal.  In the sequel, we consider the fitting of a single dimension of $\bm{f}$ with corresponding scalars $f$, $\mu_f$, $\Sigma_f$.

Here, we use a squared exponential kernel of the form
\begin{equation}
    k(\bm{x},\bm{x}') = \sigma_n^2\mathrm{exp}(-\frac{1}{2}(\bm{x}-\bm{x}')^T\mathrm{diag}(\bm{l}^{-2})(\bm{x}-\bm{x}')), \label{eq:GP_cov}
\end{equation}
with hyperparameters $\sigma_n^2$ for the noise variance and $\bm{l}\in\mathbb{R}^6$ as the length scale. The GP then defines a marginal Gaussian distribution for a test point $\bm{x}$ as
\begin{equation}
    p(f|\bm{x}, \mathcal{D}, m, k) \sim \mathcal{N}(\mu^f,\Sigma^f),
\end{equation}
where mean $\mu^f = m(\bm{x})+\bm{k}_*^T(\bm{K}+\sigma_n^2I)^{-1}[f_1,\dots,f_M]^T$, covariance $\Sigma^f = \sigma_n^2-\bm{k}_*^T(\bm{K}+\sigma_n^2I)^{-1}\bm{k}_*$ are built from kernel matrices $\bm{k}_*=[k(\bm{x}_1, \bm{x}),\dots,k(\bm{x}_M, \bm{x})]^T)\in\mathbb{R}^{M}$ and $\bm{K}\in\mathbb{R}^{M\times M}$, where the $i,j^{th}$ element is $k(\bm{x}_i, \bm{x}_j)$.

A straightforward way to incorporate prior knowledge is a mean function. A common mean function is the zero function, $m(\bm{x})=0$. To better model contact forces $\bm{f}^c$, a hinge function is also used for the mean: 
\begin{eqnarray}
    m(\bm{x}) & = &
    \begin{cases}
      c_{1}, & \text{if}\ {x}\leqq {c}_3 \\
      {c}_{1} + {c}_{2}{x} , & \text{otherwise}
    \end{cases} \label{eq:gp_hinge_mean_fkt}
\end{eqnarray}
where $x$ is the element of $\bm{x}$ which matches the dimension of the force being fit $f$, and constants ${c}_{1,2,3}\in\mathbb{R}^6$ are hyperparameters which are fit from data.

The data for these models are collected from initial demonstrations where the robot is manually guided by the human. These models are built and the hyperparameters are fit (according to negative log-likelihood) independently for each mode $n$ and dimension of $\bm{f}$. By implementing them in an automatic differentiation framework, the gradient of the mean and covariance with respect to pose can be efficiently calculated for the MPC problem.

\subsection{Robot admittance dynamics}
This subsection develops the dynamics of a robot with a Cartesian admittance controller. An admittance is used as the hardware implementation here uses a position-controlled industrial robot. As impedance and admittance are just inverse mathematical viewpoints of the interactive dynamics, the dynamics here can be readily adapted to impedance-controlled robots. Consider a rendered admittance in the TCP frame which approximates the continuous-time dynamics of
\begin{equation} \label{eq:SysDynCont}
\bm{f} - \bm{f}^r = \bm{M}\Ddot{\bm{x}} + \bm{D}\Dot{\bm{x}} + \bm{K}(\bm{x}-\bm{x}_0),
\end{equation}
with pose $\bm{x}\in\mathbb{R}^6$, velocity $\Dot{\bm{x}}\in\mathbb{R}^6$, acceleration $\Ddot{\bm{x}}\in\mathbb{R}^6$, measured force $\bm{f}\in\mathbb{R}^6$ and force reference $\bm{f}^r$. The impedance parameters are inertia $\bm{M}\in\mathbb{R}^{6\times 6}$, damping $\bm{D}\in\mathbb{R}^{6\times 6}$, and stiffness $\bm{K}\in\mathbb{R}^{6\times 6}$ matrices, which are here all diagonal, and $\bm{x}_0$ which is the rest position of the spring. The rotational elements of $\bm{x}$ and $\bm{f}$ are the angles and torques, respectively, about the three axes of the TCP frame. In the sequel, $\bm{K}$ and $\bm{x}_0$ are dropped as it was found empirically that the quasi-static tracking can be handled by the MPC adjusting $\bm{f}^r$. 

\subsubsection{Integrator}
To formulate an MPC problem, the dynamics in \eqref{eq:SysDynCont} have to be discretized. Let $T_s$ be the sample time, subscript $t$ denote the value at time $t_0 + t T_s$, and state vector $\bm{\xi} = [\bm{x}^T, \dot{\bm{x}}^T]^T$. 

A first-order explicit Euler integrator of \eqref{eq:SysDynCont} has oscillation when $1-T_sD_{i}M^{-1}_{i}<0$ \cite{haninger2021a}, where $\bullet_{i}$ denotes the $i,i$-th element of a matrix. Instead, we use the semi-implicit integration of \cite{stewart2000} on \eqref{eq:SysDynCont} to find 
\begin{equation}
\begin{aligned}\label{eq:FirstOrderDiscSysDyn}
\bm{\xi}_{t+1} &= \begin{bmatrix} \bm{I} && T_s \bm{I} \\ \bm{0} && \bm{M}\left(\bm{M}+T_s\bm{D}\right)^{-1}\end{bmatrix}
\bm{\xi}_t \\
& \qquad + \begin{bmatrix} \bm{0} \\ T_s\bm{M}^{-1} \end{bmatrix}(\bm{f}_t-\bm{f}_t^r),
\end{aligned}
\end{equation}
where $\bm{I}\in\mathbb{R}^{6\times 6}$ and $\bm{0}\in\mathbb{R}^{6\times 6}$ are the identity and zero matrix, respectively.

\subsubsection{Stochastic dynamics}
We can rewrite \eqref{eq:FirstOrderDiscSysDyn} in the form of
\begin{equation}
    \bm{\xi}_{t+1} = \bm{A}_\phi \bm{\xi}_t + \bm{B}_\phi \left(\bm{f}_t - \bm{f}^r_t\right), \label{eq:lin_dynamics}
\end{equation}
where system matrices $\bm{A}_\phi$, $\bm{B}_\phi$ depend on impedance parameters $\phi=\left\{\bm{M},\bm{D}\right\}$. Consider a Gaussian distribution over state $\bm{\xi}_t\sim\mathcal{N}(\bm{\mu}_t, \bm{\Sigma}_t)$, we find
\begin{eqnarray}
    \bm{\mu}^{+}  & = & \bm{A}_{\phi} \bm{\mu} + \bm{B}_\phi \left(\bm{\mu}^f(\bm{\mu}) - \bm{f}^r\right)  \label{eq:lin_mean}\\
    \bm{\Sigma}^+ & = & \bm{A}_\phi \bm{\Sigma} \bm{A}_\phi^T + \bm{B}_\phi \bm{\Sigma}^f(\bm{\mu}) \bm{B}_\phi^T \label{eq:lin_cov},
\end{eqnarray}
where $\bullet^+$ denotes the value of $\bullet$ at time step $t+1$. Let the vectorization of the covariance matrix be denoted $\bm{\sigma}^+ = \mathrm{vec}(\bm{\Sigma}^+)$ and similarly $\bm{\sigma}^f = \mathrm{vec}(\bm{\Sigma}^f)$. Vectorizing \eqref{eq:lin_cov} gives \cite[(520)]{petersen2008}
\begin{equation}
    \bm{\sigma}^{+} = \bm{A}_{\phi}\otimes \bm{A}_{\phi}^{T}\bm{\sigma}+\bm{B}_{\phi}\otimes \bm{B}_{\phi}^{T}\bm{\sigma}^{f} \label{eq:lin_cov_vec},
\end{equation}
where $\otimes$ is the Kronecker product.

\begin{figure}
    \centering
    \includegraphics[width=0.45\textwidth]{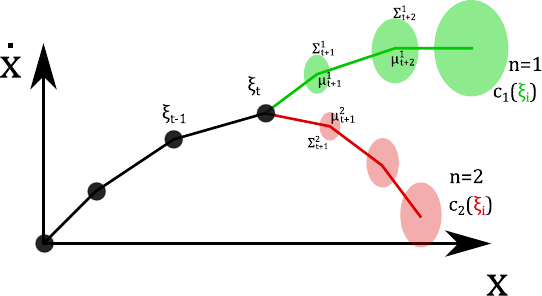}
    \caption{At time step $t$, the stochastic dynamics are rolled out from $\xi_t=[x_t, \dot{x}_t]$, with a trajectory for each mode $n$ resulting in the green and red trajectories of mean and covariance.}
    \label{fig:multimodal_traj}
\end{figure}

\subsection{Human kinematics}
Ergonomic costs are more naturally expressed in terms of human joint torques \cite{maurice2019,peternel2021}, so we also consider a 4-DOF kinematic model of the human arm with three rotational DOF at the shoulder and one rotational joint for the elbow extension. This model has parameters of $l_1$ and $l_2$, for the length of the upper and lower arm, and $\bm{x}^{sh}\in\mathbb{R}^3$ for the spatial position of the human shoulder. Using these parameters and the human joint angles $\bm{q}\in\mathbb{R}^4$, we can calculate the human hand position $\bm{x}^H\in\mathbb{R}^3$ with forward kinematics as
\begin{equation}
    \bm{x}^H =  \mathrm{FK}(\bm{q},\bm{x}^{sh}), \label{human_FK}\\
\end{equation}
Forces measured at the end-effector can be translated into human joint torques $\bm{\tau}_H$ as
\begin{equation}
    \bm{\tau}_H = \bm{J}^T_H(\bm{q})\bm{f}_H, \label{human_torque}
\end{equation}
where $\bm{f}_H\in\mathbb{R}^3$ are linear forces at the human hand, and $\bm{J}_H(q)\in\mathbb{R}^{3\times 4}$ is the standard Jacobian matrix of \eqref{human_FK} with respect to $\bm{q}$. 

Task-specific knowledge can be used to model $\bm{x}^H$ or $\bm{q}$. In most cases, we simplify the human kinematics by assuming the internal/external rotation of the arm is zero, $q_3=0$, and that $\bm{x}^H=\bm{x}$, i.e., the human hand is at the robot TCP, allowing us to write closed-form inverse kinematics and evaluate \eqref{human_torque}. 

\subsection{Inference of mode $n$}
The GPs are then used in a Bayesian estimate of the current mode, where a belief at time $t$ is denoted as $\bm{b}_t\in\mathbb{R}^N$,  $\bm{b}_t[n]=p(n|\bm{f}_{1:t}, \bm{x}_{1:t})$, and $\bullet_{1:t}=[\bullet_1,\dots,\bullet_t]$. This belief is updated recursively as
\begin{equation}
\begin{split}
    \bm{b}_{t+1}[n] \propto \bm{b}_{t}[n] \det\left(2\pi\bm{\Sigma}^f_{n}\right)\cdot \qquad\qquad\qquad\\ \qquad\qquad\qquad\exp\left((\bm{f}_t-\bm{\mu}^f_{n})^T(\bm{\Sigma}^f_{n})^{-1}(\bm{f}_t-\bm{\mu}^f_{n})\right),
\end{split}
\end{equation}
where $\bm{\mu}^f_{n}$ and $\bm{\Sigma}^f_{n}$ are the mean and covariance of the GP model in mode $n$ evaluated at $\bm{x}_t$, and $\bm{b}_t$ is normalized after the update such that $\sum \bm{b}_t[n] =1$.

\section{MPC Problem \label{sec:mpc}}
This section presents the MPC optimization problem: the constraints, cost functions, and multiple shooting formulation. This aspect is highlighted by the green part in Fig. \ref{fig:overview}.

\begin{table*}[h]
\renewcommand{\arraystretch}{1.6} 
\begin{center}
\caption{Objectives and constraints of the MPC framework, with definition, validation, and average MPC loop rate as evaluated by playing back the data recorded from the rail assembly problem.}
\begin{tabular}{r l l  l l r } 
                                                  & Feature             & Functionality   & Defined & Experiment & \makecell{Rate (Hz)} \\[0.5ex]  \hline
                                                  & baseline            & -                       & &                                       & $26.2$ \\ \cline{1-6}
 \multirow{3}{*}{\rotatebox[origin=c]{90}{Human}} & discrete modes      & human goal estimation   & & dual peg-in-hole \S\ref{sec:exp_1dof} & $10.6$  \\  \cline{2-6}
                                                  & human joint torque  & human ergonomics        & \eqref{human_torque} & polishing \S\ref{sec:exp_polish}    & $16.6$ \\  \cline{2-6}
                                                  & model variance      & repeatable task DOF     & \eqref{eq:GP_cov} & rail \S\ref{sec:exp_rail}           & $24.2$ \\ 
 \multirow{4}{*}{\rotatebox[origin=c]{90}{Task}}  &                     & constraint uncertainty  & \eqref{eq:GP_cov} & contact \S\ref{sec:exp_1dof}        & \\  \cline{2-6}
                                                  & disturbance reject  & task performance        & \eqref{eq:h2}         & polishing   \S\ref{sec:exp_polish}    & $14.6$ \\  \cline{2-6}
                                                  & well-damped         & contact stability       & \eqref{eq:damped_con} & contact     \S\ref{sec:exp_1dof}      & $22.2$ \\  \cline{2-6}
                                                  & chance constraint   & safety                  & \eqref{eq:chance_con} & contact     \S\ref{sec:exp_1dof}      & $13.0$ \\  \hline
                                                 
 \hline
\end{tabular}
\label{tab:mpc_problems}
\end{center}
\end{table*}

\subsection{Optimization Problem \label{sec:opt_prob}}
The MPC problem is formulated using multiple-shooting transcription with a generic problem statement of
\begin{equation}
\begin{array}{r@{}ll}
    \bm{u}_{t:t+H} = & & \S\ref{sec:opt_prob} \\
     &{ \arg\min_{\bm{u}} \sum_{t}^{t+H}\sum_{n=1}^N b_t[n] c_n(\bm{\xi}^n_t, \bm{u}_t) }& \S\ref{sec:costs}\\
    \mathtt{s.t.} \,\,\,\, &\forall n \in [1,\dots,N], \tau \in [t,\dots,t+H-1]: &\\
    &{\bm{\mu}^n_t = \bm{\xi}_t,\,\,\bm{\Sigma}^n_t = 0} &\\
    &{\vert\bm{\mu}^n_{\tau+1}-f(\bm{\mu}^n_{\tau},\bm{u}_\tau)\vert \leq \varrho} & \eqref{eq:lin_mean} \\
    &{\vert\bm{\Sigma}_{\tau+1}-g(\bm{\mu}^n_{\tau},\bm{\Sigma}^n_{\tau},\bm{u}_\tau)\vert \leq \varrho} & \eqref{eq:lin_cov_vec} \\
    &{\bm{h}\left(\bm{\xi}_t,\bm{u}_t\right) \geq 0} & \S\ref{sec:constraints}\\ 
    &{\bm{u} \in U}
    \label{mpc_statement}
\end{array}
\end{equation}
where $H$ is the planning horizon, $U$ is the range of allowed inputs, $\varrho$ the slack for the continuity constraints (the inequality is applied element-wise), $f$ is from \eqref{eq:lin_mean}, $g$ following \eqref{eq:lin_cov_vec}, and $\bm{h}$ are inequality constraints to be presented in \ref{sec:constraints}. The cost function $c_n$ is shown with it's parameter argument $\xi_t$ and decision variable $\bm{u}$, the detailed cost function provided in \eqref{sec:cost_fn}. Initial covariance is set $\Sigma_t = 0$ as it's assumed that the robot state is fully observed.

The constraints in \eqref{mpc_statement} are nonlinear, so an interior-point nonlinear optimization solver is used. While nonlinear, the problem is writtein in an automatic differentiation framework, allowing calculation of the gradient and Hessian of the objective and constraints, significantly improving convergence.

The MPC framework here allows for different choices of decision variable $\bm{u}$, which can include any of the following:
\begin{eqnarray}
\bm{f}^r_{t:t+H} & & \mathrm{Robot\,\,trajectory} \nonumber\\
\bm{\Delta}^M_t, \bm{\Delta}^D_t & & \mathrm{Change\,\,in\,\,robot\,\,impedance} \nonumber
\end{eqnarray}
where $\bm{\Delta}^M_t$ and $\bm{\Delta}^D_t$ are a change in the impedance parameters $\bm{M}$ and $\bm{D}$.

\subsection{Costs \label{sec:costs}}

\subsubsection{Stage cost function $c_n$ \label{sec:cost_fn}}
The general stage cost function is defined as
\begin{equation}
\begin{split}
    c_n(\bm{\mu}, \bm{\Sigma}, \bm{\mu}^{f,n}, \bm{\Sigma}^{f,n}, \bm{\tau}_H, \bm{u}) & = \\
      & \bm{\mu}^T\bm{Q}_\mu\bm{\mu} + \mathrm{Tr}(\bm{Q}_\Sigma{\bm{\Sigma}}) + \\ 
      &  (\bm{\mu}^{f,n})^T\bm{Q}_f\bm{\mu}^{f,n} + \mathrm{Tr}(\bm{Q}_{\Sigma,f}\bm{\Sigma}^{f,n})  \\
          &  + \bm{\tau}_H^T\bm{Q}_\tau\bm{\tau}_H + {\bm{u}}^T\bm{Q}_u\bm{u},
\end{split} \label{cost_fn}
\end{equation}
where $\bm{\mu}$ and $\bm{\Sigma}$ are mean and covariance for the predicted state $\bm{\xi}$ in mode $n$, $\bm{\mu}^{f,n}$ and $\bm{\Sigma}^{f,n}$ are the mean and covariance of the GP forces in mode $n$, $\bm{\tau}_H$ are the human joint torques from \eqref{human_torque}, $\bm{u}$ are control inputs, and $\bm{Q}_{\cdot}$ are the related weight-matrices for each cost factor. State cost $\bm{Q}_\mu$ includes just a small penalty on velocity, the trajectory following is implicit in GP mean and covariance. In some applications (e.g. the rail assembly here \ref{fig:contact_modeling}), replacing $\left(\bm{\mu}^{f,n}\right)^T\bm{Q}_H\bm{\mu}^{f,n}$ with $(\bm{\mu}^{f,n}+\bm{f}^r_t)^T\bm{Q}_H(\bm{\mu}^{f,n}+\bm{f}^r_t)$ is used to match the human demonstrations.   

\subsubsection{Human Joint Torque Cost}
Human arm manipulability plays a key role in ergonomics \cite{gopinathan2017,peternel2021}, as it relays the information about how well the joint torques can be transferred into the end-effector force where the task is produced. Therefore, we defined the ergonomic cost as
\begin{eqnarray}
    \bm{\tau}_H^T\bm{Q}_\tau\bm{\tau}_H & = \bm{f}_H^T\bm{J}_H(\bm{q})\bm{Q}_\tau\bm{J}_H(\bm{q})^T\bm{f}_H,
\end{eqnarray}
where, when $\bm{Q}_\tau=\bm{I}$, $\bm{J}_H\bm{J}_H^T$ defines the standard manipulability ellipsoid. This is a general formulation and the composition of the human end-effector force $\bm{f}_H$ can be specified based on the task. For example, in the case of a polishing task, the structure to be polished itself supports the gravity force, which thus does not play a key role. Here, the dominant force component is related to the movement of the robot along a path, therefore we use the damping forces of the robot $\bm{f}_H=\bm{D}\dot{\bm{x}}$, where $\dot{\bm{x}}$ provides a weighting according to the direction of the expected forces. This term explicitly includes the impedance parameter, allowing the damping to be optimized to direction of expected motion and ergonomics. Nevertheless, the proposed formulation is general and $\bm{f}_H$ can include gravity force and other static forces for tasks such as carrying and drilling.

\subsubsection{Disturbance Rejection}
In many tasks, certain forces are \textit{external perturbations} and should not influence the robot position, e.g., forces arising from the rotation of a polishing disk in a polishing task. In some cases, these forces may be more easily distinguished in the frequency domain, where typical human forces are low-frequency, and disturbance forces may be medium- to high-frequency. Frequency-dependent costs can be considered through an augmented state model \cite{grandia2019}, but this increases state dimension (and thus computational complexity) and requires the cost can be expressed from the inputs or state. Here, measured force is not an input to the MPC problem, thus a new approach is taken.

We consider a disturbance force model in Laplace domain as $F^d(s) = \frac{\alpha s}{s+\omega_p}$, where $\alpha$ and $\omega_p$ adjust the magnitude and cut-off frequency. This model is a simple high-pass filter, distinguishing from lower-frequency human input. This disturbance results in robot position deviation according to the admittance dynamics $X^d(s) = A(s)F^d(s)$, where $A(s) = (M_is^2+D_is+K_i)^{-1}$. Recall the equivalency of the two-norm in time and Laplace domain, $\Vert x^d(t) \Vert^2_2= \Vert X^d(s) \Vert^2_2$ by Parseval's theorem \cite{zhou1996}, and that $\Vert X^d(s) \Vert_2^2$ can be found by the observability grammian as 
\begin{eqnarray}
\Vert X^d(s) \Vert^2_2 & = & B_d^T X_o B_d, \label{eq:h2} \\
\mathrm{where}& & A_dX_o +X_oA_d^T = C_d^TC_d, \label{eq:h2_lyap}
\end{eqnarray}
where $A_d, B_d, C_d$ are a state-space realization of $X^d(s)$ \cite{zhou1996}. The Lyapunov equation in \eqref{eq:h2_lyap} is differentiable \cite{kao2020} with respect to admittance parameters $M_i, D_i, K_i$, allowing the efficient calculation of the Jacobian and Hessian of this objective for the optimization problem. Note that the na\"ive solution of \eqref{eq:h2_lyap}, by vectorizing and a linear solve may have numerical issues when the matrices are poorly conditioned, in such case using specialized solvers like Bartels-Stewart can improve performance. 

\subsection{Constraints \label{sec:constraints}}
Impedance parameters affect contact safety.  To improve contact safety, constraints for well-damped contact and force limit are formulated.

\subsubsection{Force limit chance constraint}
A fundamental concern in contact is avoiding excessive force; with a model of force over robot pose, this constraint can be directly transcribed. This force model may have uncertainty arising from:
\begin{enumerate}
    \item Variation in  \textit{environment constraint} location
    \item Exploration of a new region or \textit{confidence in the skill}
\end{enumerate}
Both these uncertainties are captured in the GP regression here. For 1), the GP will fit hyperparameters which increase the uncertainty in the force model (validated in \S\ref{sec:analysis_contact}). For 2), when farther from the training data, the GP will revert to a baseline variance given by hyperparameter $\sigma_n^2$. 

A chance constraint is a constraint applied to a random variable which holds for a specified degree of certainty. For a scalar Gaussian variable $f \sim \mathcal{N}(\mu^f, \Sigma^f)$, a chance constraint can be re-written over mean and covariance as
\begin{equation}
    \mathrm{Prob}\{f \leq \overline{f} \} > 1-\epsilon  \iff  \mu^f < \overline{f} - \mathrm{erf}^{-1}(1-\epsilon)\Vert\sqrt{\Sigma^{f}}\Vert , \label{eq:chance_con}
\end{equation}
where $\overline{f}$ is an upper bound on the force, $\mathrm{erf}^{-1}$ is the inverse of the error function and $\epsilon$ is a tuning parameter for the confidence that the chance constraint should hold \cite{hewing2019}. 

It can be seen that the uncertainty adds a conservatism to the force bound; as $\Sigma^f$ gets larger, the mean force $\mu^f$ must be kept smaller. This constraint is applied element-wise on the linear forces, evaluating the GP at the final MPC trajectory point $\bm{\xi}_{t+H}$ to reduce computational complexity.

\subsubsection{Well-damped contact constraint}
One challenge in admittance control is contact stability with \textit{environment constraints}, where a higher-stiffness environment typically requires higher virtual damping \cite{albu-schaffer2003}. Ad hoc methods can increase damping near contact \cite{ficuciello2015}, but make assumptions such as the robot is slowing down before contact. Here, the environment stiffness can be estimated from the mean function of the GP model as
\begin{eqnarray}
    K_{e,i}(\bm{x}_t) \approx \left.\frac{d \mu^f_i(\bm{x})}{dx_i}\right|_{\bm{x}=\bm{x_t}}, \label{eq:env_stiff}
\end{eqnarray}
where $K_{e,i}$ is the estimated environment stiffness in the $i^{\mathrm{th}}$ direction at pose $\bm{x}_t$. With this estimated stiffness, the dynamic parameters can be directly constrained using the condition for well-damped 2nd order systems
\begin{eqnarray} 
    D_i \geq 2\zeta\sqrt{M_iK_{e,i}(\bm{x}_t)}, \label{eq:damped_con}
\end{eqnarray}
where $\zeta = 1$ is critically-damped, and $\zeta>1$ introduces a safety margin which can help account for the non-idealized admittance dynamics. This constraint is applied to the entire MPC trajectory $\bm{\xi}_{t+1:t+H}$.

\section{Analysis\label{sec:analysis}}
This section, shown in purple in Fig. \ref{fig:overview} shows some properties of the MPC problem: how trajectory and impedance optimization differ, necessary conditions for convexity, and investigates the ability to model contact with GPs.

\subsection{Optimization of Impedance vs Trajectory}
Why should we optimize impedance parameters $\phi$ instead of just the trajectory $\bm{\mu_}\bullet$? Taking the differential of the linear-Gaussian dynamics \eqref{eq:lin_mean} and \eqref{eq:lin_cov_vec},
\begin{multline}
 \left[\begin{array}{c}
        d\bm{\mu}^+ \\ d\bm{\sigma}^+ 
  \end{array}\right] = \\
  \left[\begin{array}{c c}
        \bm{B}_\phi & d_\phi\bm{A}_\phi\bm{\mu} + d_\phi\bm{B}_\phi \bm{\mu}^f \\ 
        0 & d_\phi(\bm{A}_\phi \otimes \bm{A}_\phi^T) \bm{\sigma} + d_\phi(\bm{B}_\phi \otimes \bm{B}_\phi^T) \bm{\sigma}^f
  \end{array}\right] \left[\begin{array}{c}
        d\bm{f}^r \\ d\phi 
  \end{array}\right],\label{eq:differential}
\end{multline}
where prefix $d$ indicates total differential and $d_\bullet$ the partial derivative with respect to $\bullet$. We note two properties: i) input $\bm{f}^r$ doesn't affect the covariance $\bm{\sigma}^+$, and ii) state $\bm{\mu}^+$ is determined by both control input $\bm{f}^r$ and impedance parameters $\phi$, i.e., different combinations of force and impedance parameters can realize a certain change in $\bm{\mu}^+$, as previously noted in \cite{stulp2012}. 

When the robot increases stiffness, the mean and variance of the robot's trajectory are affected, for example reducing response to force disturbances. However, changing the robot's force reference only affects the mean trajectory. This shows that explicit consideration of trajectory covariance is important for impedance, thus the MPC trajectories should be stochastic when optimizing impedance. 

\subsection{Convexity over impedance parameters $\phi$}
For the optimization problem to be efficient, the objective should be convex with respect to the decision variables \cite{rawlings2017}. As optimizing with respect to parameters of the dynamics (e.g., impedance parameters) is not studied, we check the convexity of a toy cost function with respect to impedance parameters, denoted $\phi= [\bm{M}, \bm{D}]$. As impedance parameters $\phi$ mostly affect covariance, we consider a simplified objective in a single step shooting problem of $\min_\phi\mathrm{Tr}(\bm{Q}\bm{\Sigma}^+)$, where by results from \cite{petersen2008}, 
\begin{eqnarray}
    \mathrm{Tr}(\bm{Q}\bm{\Sigma}^+) & = & \mathrm{Tr}(\bm{Q}\bm{A}_\phi \bm{\Sigma} \bm{A}_\phi^T + \bm{Q}\bm{B}_\phi \bm{\Sigma}^f \bm{B}_\phi^T) \label{eq:trace_obj} \\
    & = & \bm{\sigma}^T\mathrm{vec}(\bm{A^TQA})+(\bm{\sigma}^f)^T\mathrm{vec}(\bm{B^TQB})\nonumber \\
    & = & \left(\bm{\sigma}^T(\bm{A}^T\mathtt{\otimes}\bm{A}^T)\mathtt{+}(\bm{\sigma}^f)^T(\bm{B}^T\mathtt{\otimes}\bm{B}^T)\right)\mathrm{vec}(\bm{Q}).\nonumber
\end{eqnarray}
The convexity of this objective will depend on the impedance parameter values, the choice of the integrator (which affects $\bm{A}$ and $\bm{B}$), and the relative magnitude of $\bm{\sigma}$ and $\bm{\sigma}^F$. Note that $\bm{B}^T\otimes \bm{B}^T = [0,0,0,T_s\bm{M}^{-2}]$, which gives $\nabla_{\bm{M}}^2 \bm{B}^T\otimes \bm{B}^T \succeq 0$. Thus increasing $\bm{\Sigma}^f$ can increase the minimum eigenvalue of the Hessian of the objective \eqref{eq:trace_obj}. The terms for damping are more complex, thus numerical analysis is carried out. 

\begin{figure}
    \centering
    \includegraphics[width=0.90\columnwidth]{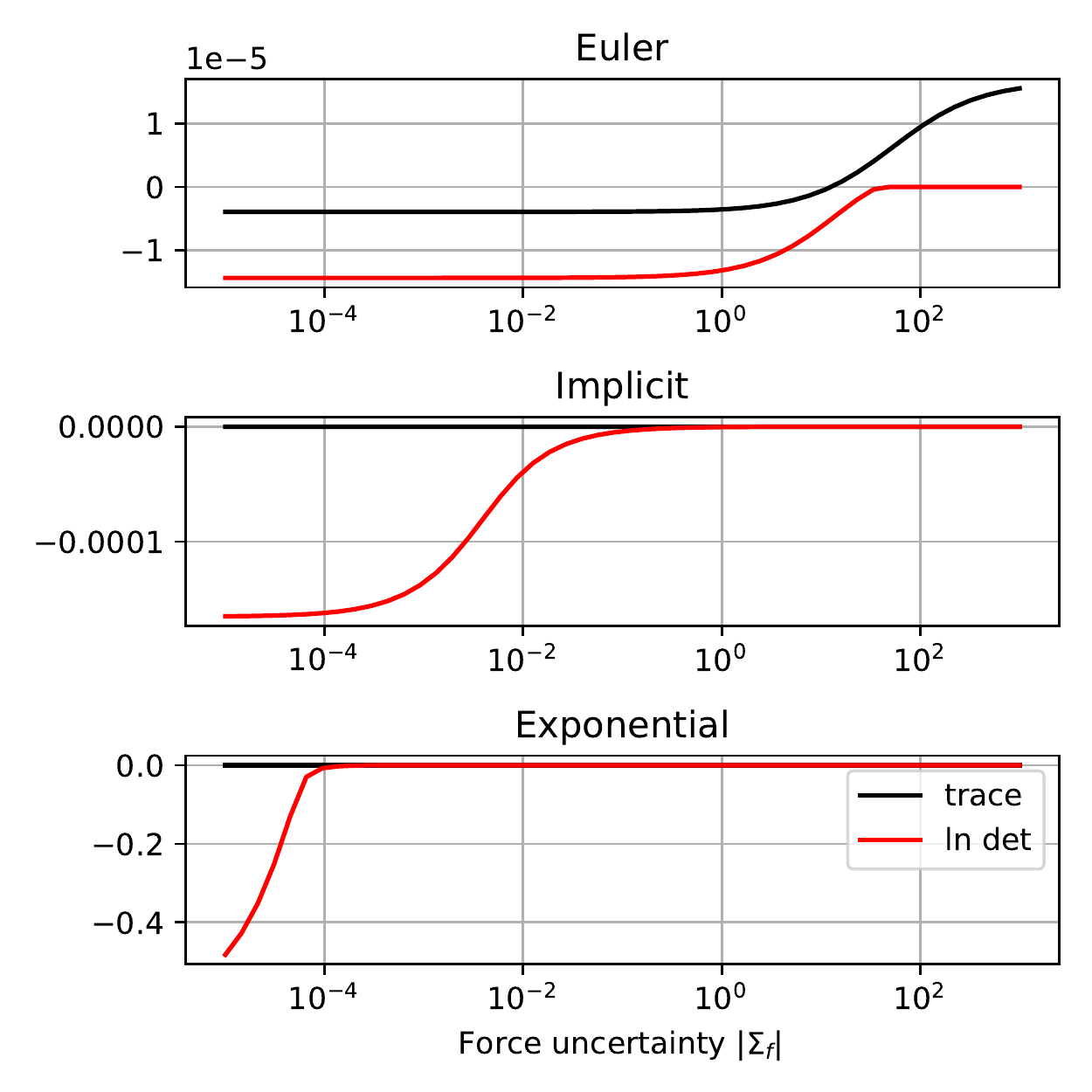}
    \caption{Minimum eigenvalue of the Hessian $\nabla^2 \mathrm{Tr} \bm{\Sigma}^+$ and $\nabla^2 \ln\det\bm{\Sigma}^+$ with respect to impedance parameters as $\bm{\Sigma}_f$ varies, where the minimum eigenvalue must be positive for this objective to be convex. The integrator schemes (Euler/implicit/exponential) and objective ($\mathrm{Tr}$ vs $\ln\det$) affect the convexity, where the implicit and exponential integrator with trace performs best over a range of force uncertainty.}
    \label{fig:hessian}
\end{figure}

These results are validated numerically to see what matrix cost and integrator have better convexity. The minimum eigenvalue of the Hessian $\nabla^2_\phi\mathrm{Tr}(\bm{Q}\bm{\Sigma}^+)$ and $\nabla^2_\phi\ln\det(\bm{Q}\bm{\Sigma}^+)$ are compared for three different integrator schemes: explicit Euler $\dot{x}^+=(1+T_sD/M)\dot{x}$, implicit $\dot{x}^+=M(M+T_sD)^{-1}\dot{x}$, and exponential $\dot{x}^+=\exp(-T_sD/M)\dot{x}$. We can see in Fig. \ref{fig:hessian} that the minimum eigenvalue generally increases as $|\bm{\Sigma}^f|$ increases. Using a $\mathrm{Tr}(\bm{Q}\bm{\Sigma}^+)$ is unilaterally better than $\ln\det(\bm{Q}\bm{\Sigma}^+)$, and is thus used in the cost function here. The implicit integrator has better average performance over the range of $\bm{\Sigma}^f$---notably, the performance of the explicit Euler integrator is poor; the Hessian is indefinite for typical parameter values. Thus, here we use the trace of matrix costs with an implicit integrator. Also note that convexity improves as $\bm{\Sigma}^f$ increases, as suggested in the above analysis.

\subsection{Contact modeling \label{sec:analysis_contact}}
We investigated the ability of the GPs to model contact by verifying 1) that GPs can identify the contact in human-guided contact scenarios, 2) the impact of variation in the contact position, and 3) the impact of environment stiffness on the ability to regress the models. These together allow the direct modeling of \textit{environment constraint} uncertainty in realistic contact. The hyperparameters in the following were all fit to minimize negative log-likelihood from the same initial values. 

\begin{figure}[t]
    \centering
    \subfloat[Human vs autonomous contact\label{fig:contact_auto}]{\includegraphics[trim={0 12 0 10},clip,width=0.75\columnwidth]{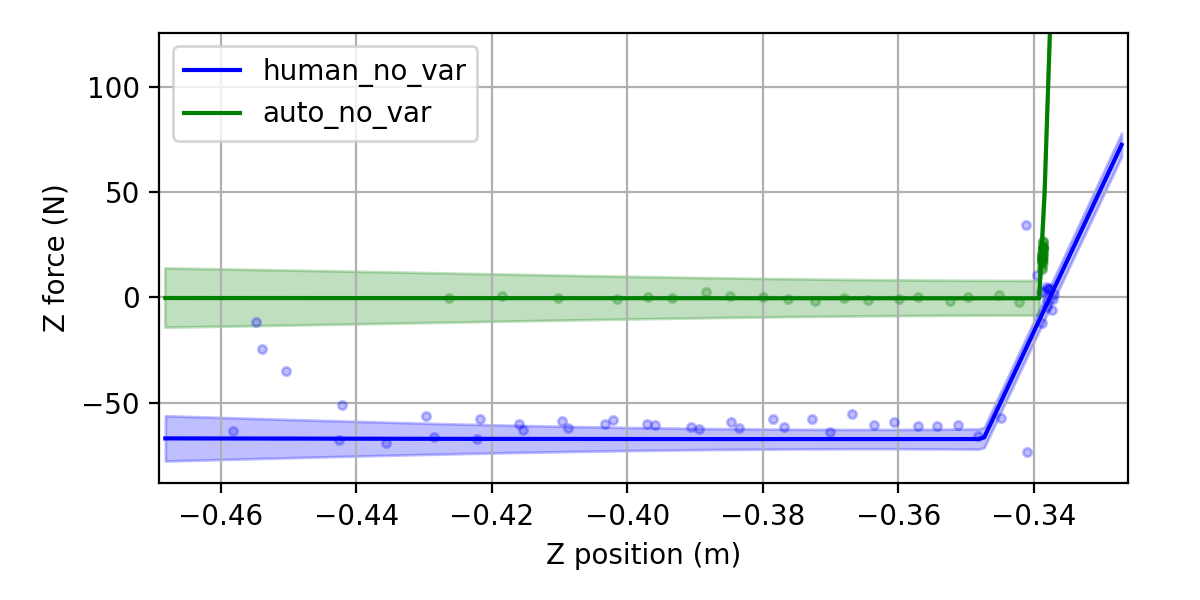}} \\
    \subfloat[Increased variation in contact position\label{fig:contact_var}]{\includegraphics[trim={0 12 0 10},clip,width=0.7\columnwidth]{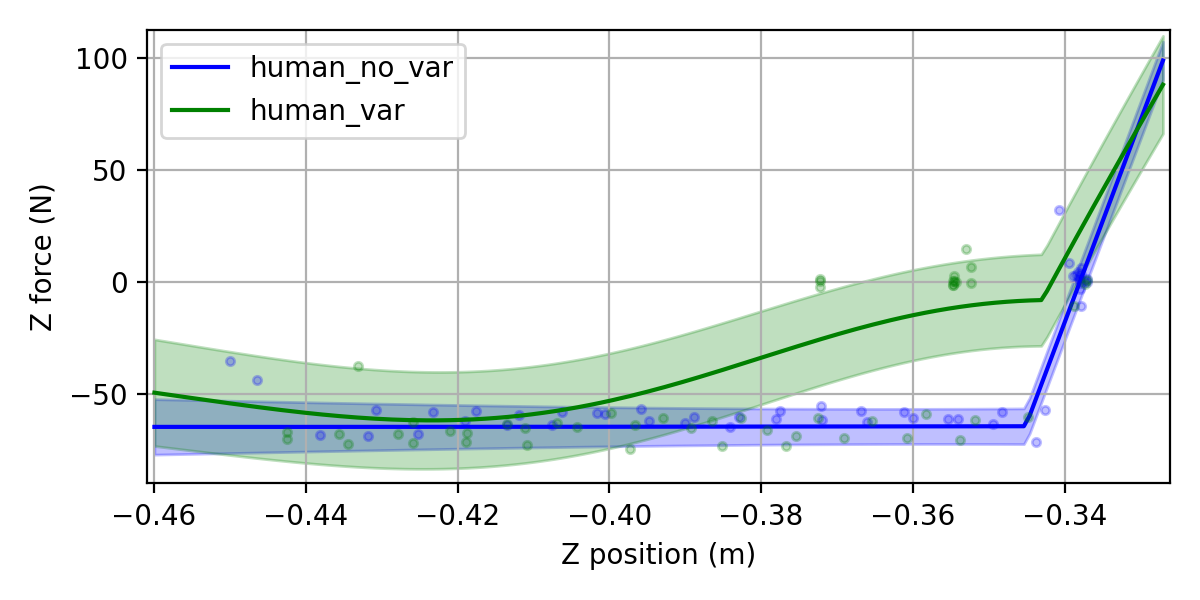}} \\
    \subfloat[Changes in stiffness and contact location, hinge mean \label{fig:contact_stiff}]{\includegraphics[trim={0 12 0 10},clip,width=0.7\columnwidth]{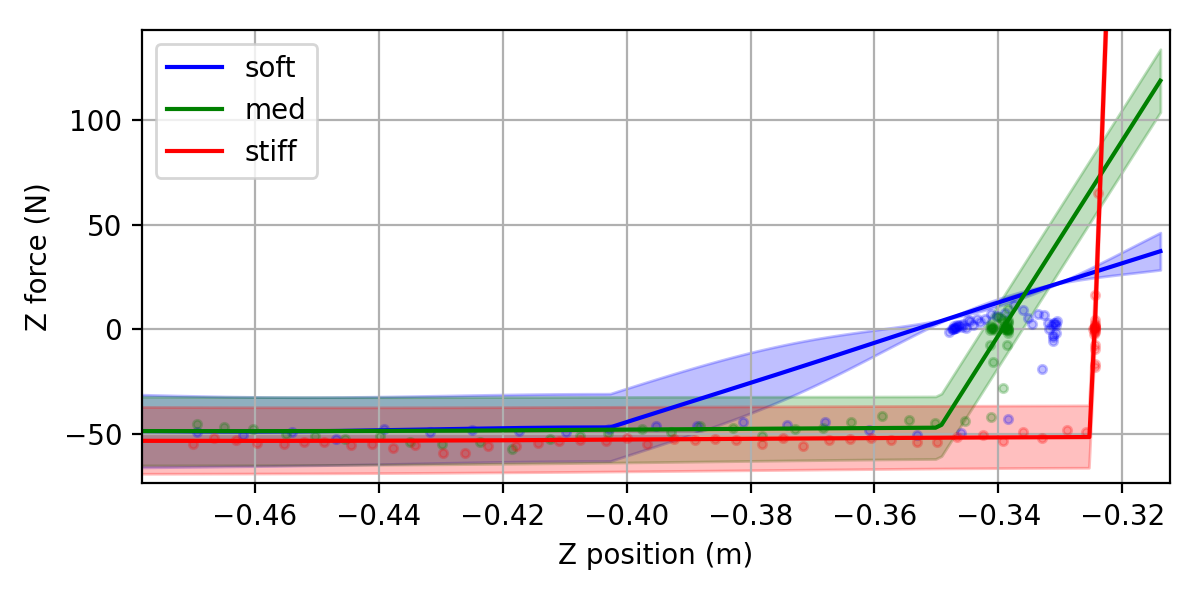}}  

    \caption{Mean and covariance of GP models for various contact situations where contact occurs on the right side at $-0.34$ m, real data in dots and corresponding model in matching color. Learning from human-guided contact (a) introduces acceptable error in contact location and stiffness, (b) variation in contact location induces higher variance GP models, and (c) the hinge function can represent a range of environment stiffnesses.}
    \label{fig:contact_modeling}
\end{figure}

When the human guides the robot into contact, the sum of human and environmental forces is measured. We collect human-guided contact and autonomous contact data (i.e. the robot makes contact without human forces), and compare the GPs which are fit to both of them. In Fig. \ref{fig:contact_auto}, we see that both models identify the contact location with the hinge mean function, but the human data identifies a lower stiffness and a contact position $1$ cm too high. In the applications here, this error was acceptable, but this is application-specific.

Next, the contact position was varied between trials, with a total variation of $2$ cm over three trials. The result can be seen in Fig. \ref{fig:contact_var}, where the three contact positions can be seen and the higher covariance in the fit GP (green) can be seen compared to the constant contact position (blue). 

Finally, contact with environments of various stiffnesses was made, with corresponding stiffness of $[15, 45, 126]$ N/mm. The resulting quality of the fit can be seen in Fig. \ref{fig:contact_stiff}, where the hinge function can readily identify the changes in contact position and stiffness. 

\section{Validation\label{sec:validation}}
This section presents four sets of experiments to validate the various features of the proposed framework. These validation experiments should provide useful pointers for the application of the proposed method in real-world scenarios (the blue part in Fig. \ref{fig:overview}). Videos are attached as multimedia and can be found at \url{https://youtu.be/Of2O3mHfM94}. MPC and GP parameters used for each experiment can be found at the repository \url{https://gitlab.cc-asp.fraunhofer.de/hanikevi/gp-mpc-impedance-control}

\subsection{Contact uncertainty and stiffness \label{sec:exp_1dof}}
We first used a contact task to validate safe contact with the environment. In particular, we examined the effects of force chance constraint and well-damped constraint as contact variance or stiffness increases. The robot was taught a task moving from free space into contact, which resulted in GP models as seen in Section \ref{sec:analysis_contact}. The contact location could be varied by adding blocks (Fig. \ref{fig:contact_var}), and the environment made stiff by locking out a compliant element (Fig. \ref{fig:contact_stiff}). For each contact condition--soft, high variance, and stiff--the corresponding model from Section \ref{sec:analysis_contact} was used, and only the MPC parameters relating to the chance constraint ($\overline{f}=12$, $\epsilon=0.5$, chosen by hand tuning) and well-damped constraint ($\xi=1.2$) are changed. The low impedance test condition kept the minimum mass $5$ Kg and damping $500$ Ns/m. 

\begin{figure}[t]
    \centering
    \includegraphics[width=0.5\textwidth]{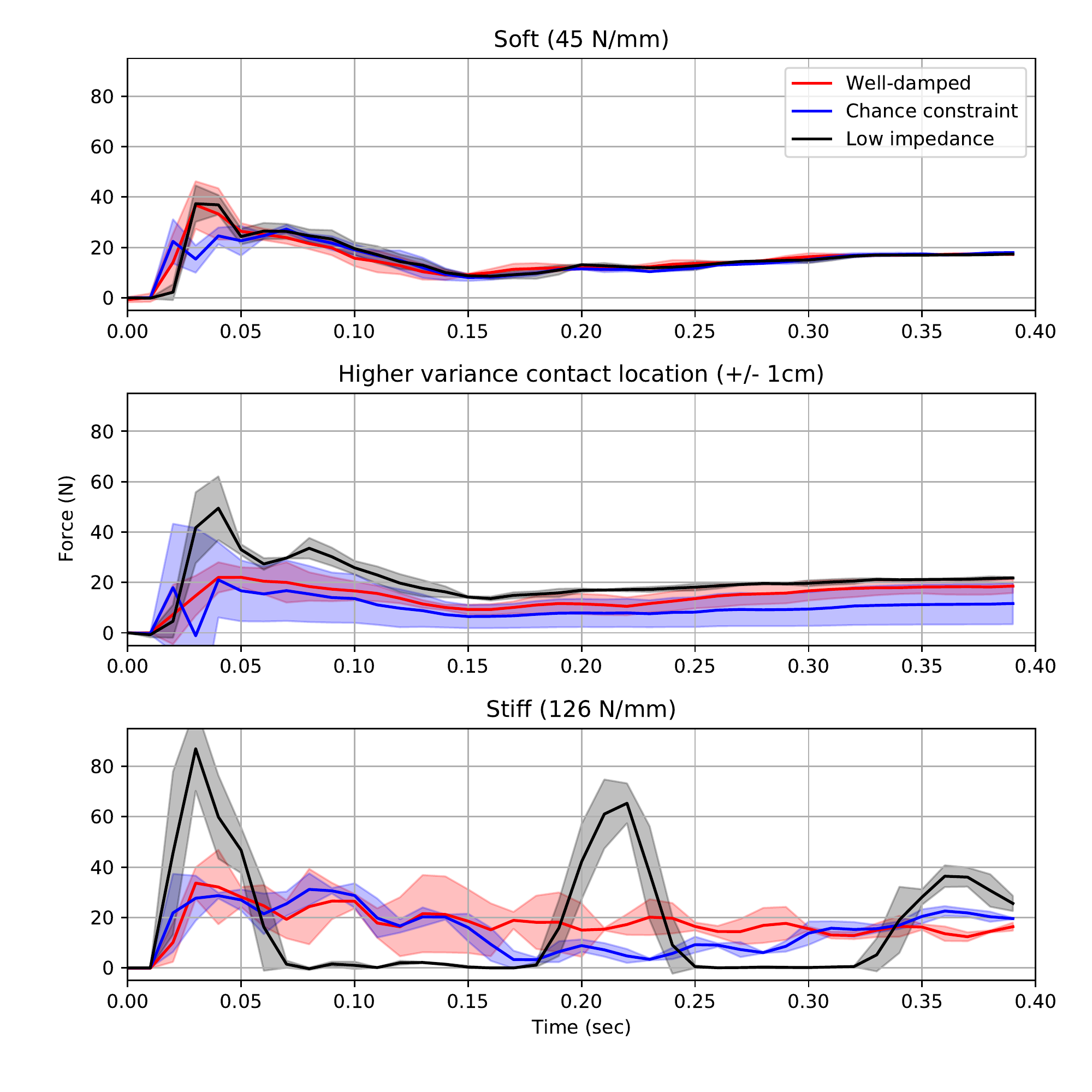} 
    \caption{Contact forces under contact conditions of soft (top, 45 N/mm), higher variance (middle, $\pm 1$ cm) and stiff (bottom, 126 N/mm). The two contact constraints are compared, chance constraint \eqref{eq:chance_con} and well-damped \eqref{eq:damped_con}. When the contact location has higher variance, the chance constraint is more conservative (middle). The well-damped constraint stabilizes contact with the high-stiffness environment by increasing damping (bottom), but does not interfere with the low-stiffness contact where increased damping is not needed (top).}
    \label{fig:exp_1dof}
\end{figure}

The force trajectories during contact are shown in Fig. \ref{fig:exp_1dof}, averaged over three trials, where variance is indicated by the shaded area. In the soft contact condition, the difference is small as the constraints were not violated. As the variance in the environment increased, the low impedance MPC had a higher contact force, but both constraints reduced this peak force. The chance constraint had more oscillation, but the well-damped constraint was smoother. As the stiffness of the environment increased, the low impepdance MPC had a higher peak contact force, and had unstable contact (i.e., lost contact due to oscillation). The well-damped constraint had more stable contact with fewer oscillations, compared to the chance-constrained MPC.  

In the supplementary video, it can be seen that these constraints also affected the behavior approaching contact. Once the chance constraint was activated, the robot pulled back from the contact approach. This was typically accompanied by infeasible optimization problems (i.e., no solution meeting constraints could be found). This problem was empirically more pronounced when using a GP with a hinge function, thus the poor conditioning of stiff contact was identified as a contributing cause. Alternatively, the well-damped constraint paused to allow the damping to increase before making contact. Also, the well-damped constraint was found empirically to have lower computational cost, with an average MPC rate of $22$ compared with $13$ for the chance constraint (Table \ref{tab:mpc_problems}).

\subsection{Rail Assembly \label{sec:exp_rail}}
Here we present a rail assembly task, where the location of the switch along the rail may vary according to factors not known to the robot, such as existing parts. However, the rail mounting itself is repeatable. This task was taught in three demonstrations from various initial conditions, with varying goal locations along the rail. The desired behavior is that this division of repeatable task DOF and variation can be automatically extracted from data, such that the robot allows the human to more easily manipulate this DOF, i.e., has a lower impedance in the DOF that varies.

\begin{figure}[t]
    \centering
    \subfloat[Rail assembly \label{fig:exp_rail_setup}]{\includegraphics[width=0.248\textwidth]{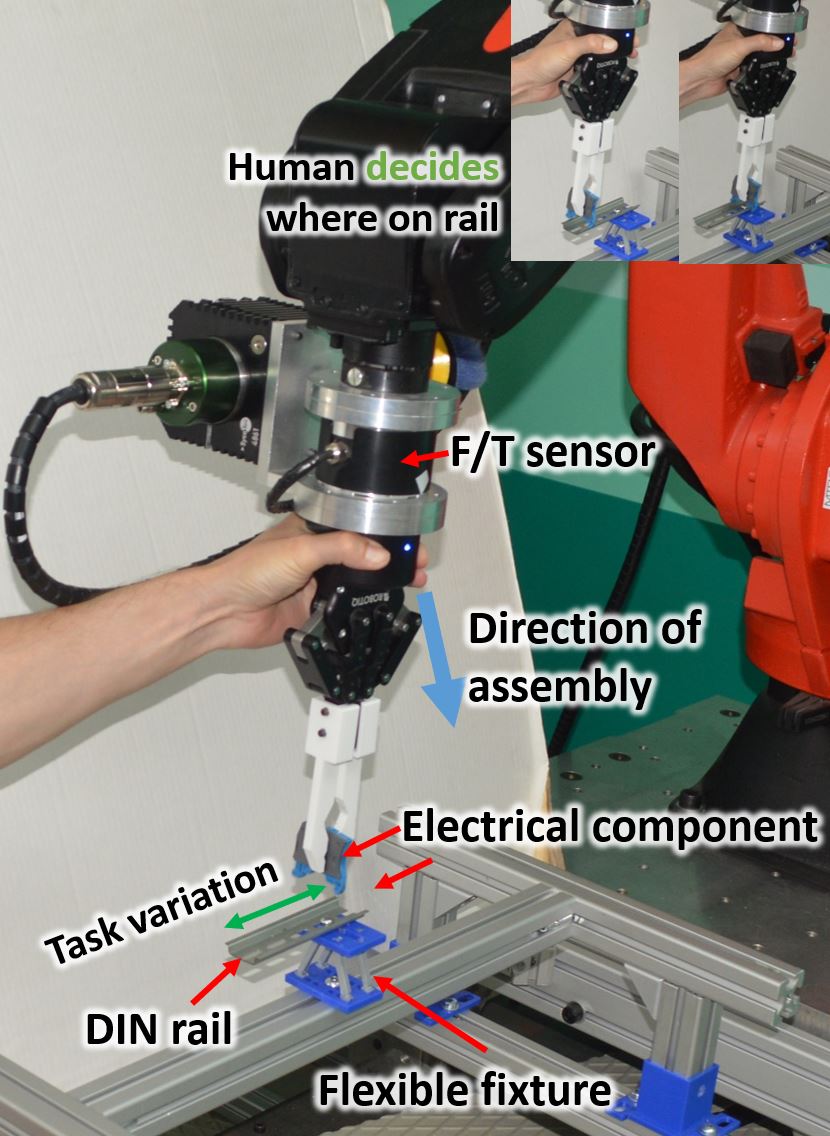}}\hfill 
    \subfloat[Polishing \label{fig:exp_polish_setup}]{\includegraphics[width=0.225\textwidth]{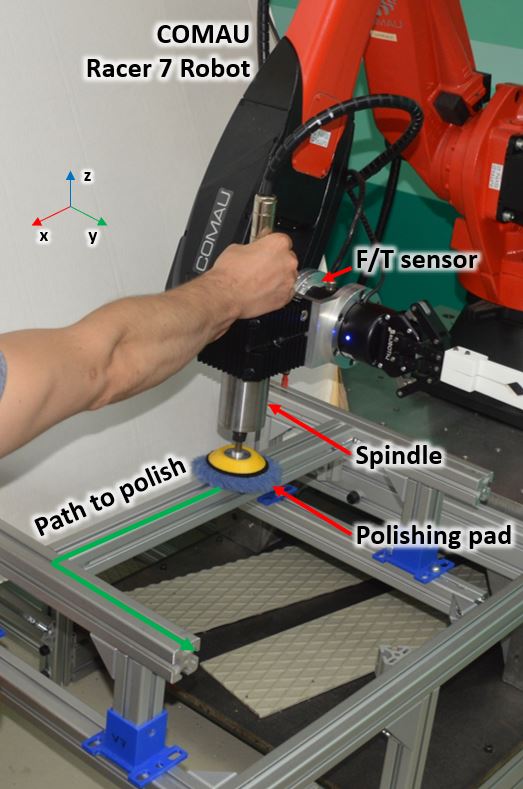}}
    \caption{Rail assembly and polishing tasks, where the DOF which the human should determine are shown in green.}
\end{figure}

\begin{figure}[t]
    \centering
    \includegraphics[trim={0 0 0 30},clip,width=0.35\textwidth]{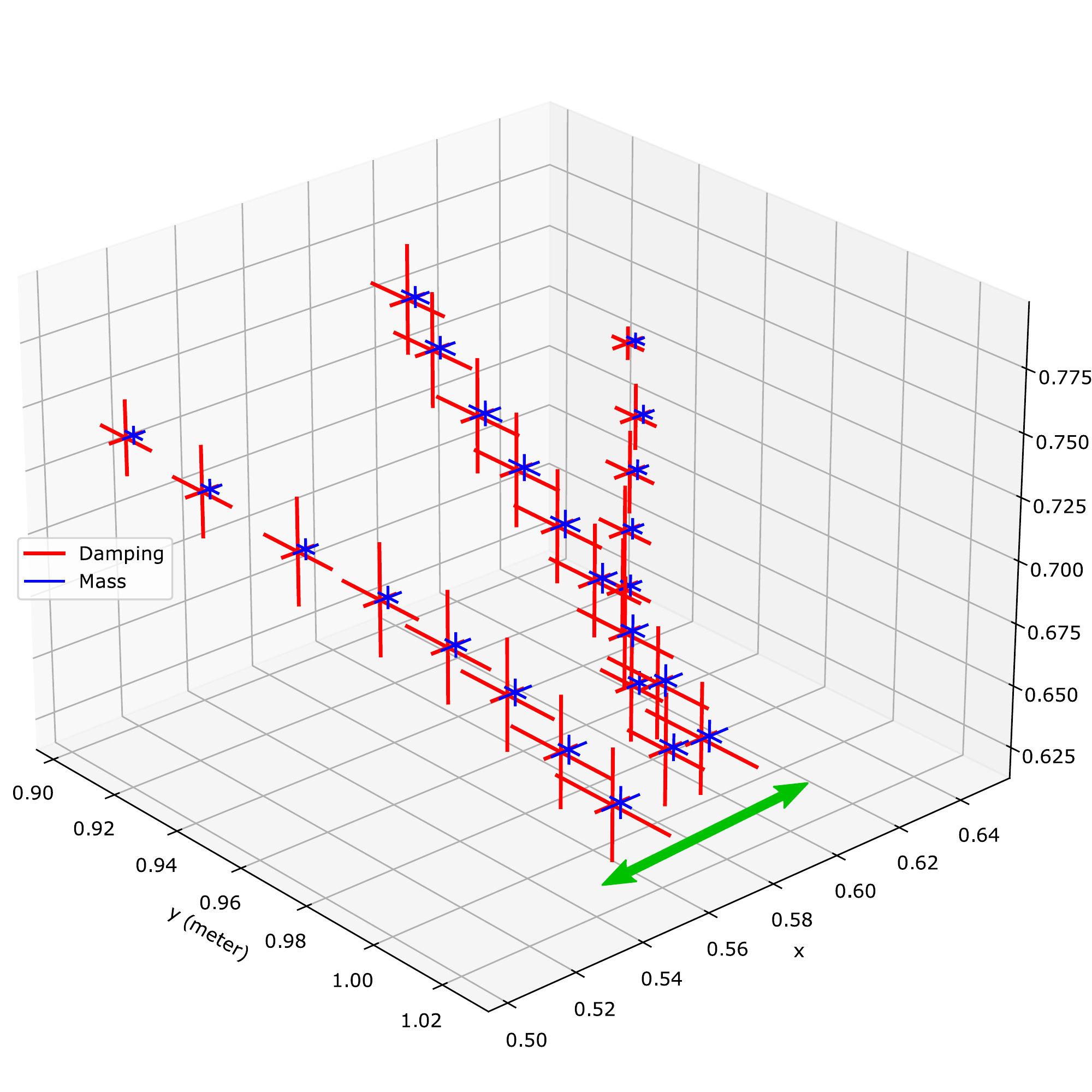}
    \caption{Impedance values in the rail assembly task, showing principle axes of damping and mass in task space along three trajectories. The DOF with task variation (green arrow) has consistently lower damping values, allowing easier human manipulation along this axis.}
    \label{fig:exp_rail}
\end{figure}

The results can be seen in Fig. \ref{fig:exp_rail}, which shows the linear impedance parameters of mass and damping over space in three validation experiments. It can be seen that the impedance initialized at low values, but as the rail was approached, the impedance in the repeatable DOF was increased substantially, while kept low in the DOF which varied. The supplementary video shows that this results in a human easily intervening to adjust position along the rail. The task could also be achieved autonomously after the demonstrations when a cost was added to keep a positive force reference in the approach direction $\Vert 15-\bm{f}^r_2\Vert$. 

\subsection{Polishing \label{sec:exp_polish}}
We also examined a polishing task where the human had to guide a robot-supported polishing spindle along a path. In this case, the human was determining speed along the path, while polishing vibration should not affect the robot's trajectory. To balance the rejection of polishing force disturbance against the need for the human to easily manipulate the robot, a force disturbance cost was added to the MPC problem. An ergonomic cost is also added, based on the assumed human arm geometry, and tested. The supplementary video shows polishing on a flat surface, corresponding to Fig. \ref{fig:exp_polish_setup}, and a round object to show the ability to handle orientation.

\begin{figure}[t]
    \centering
    \subfloat[Baseline]{\includegraphics[trim={100 80 100 80},clip,width=0.4\textwidth]{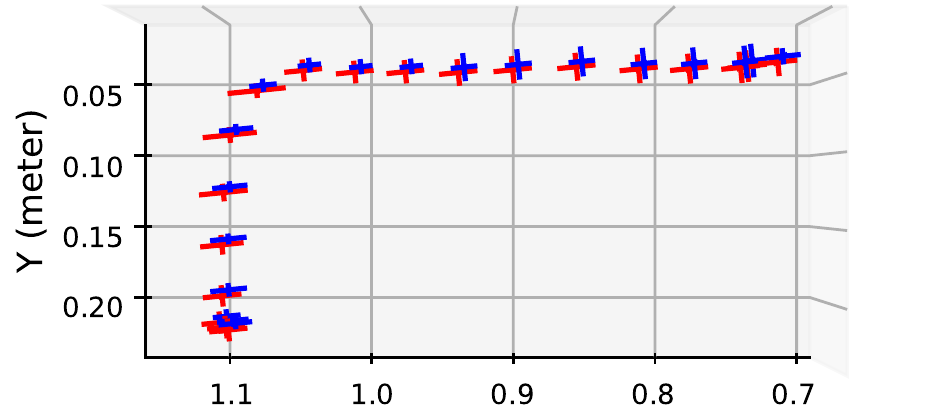}} \\ 
    \subfloat[With freq. disturbance]{\includegraphics[trim={100 80 100 80},clip,width=0.4\textwidth]{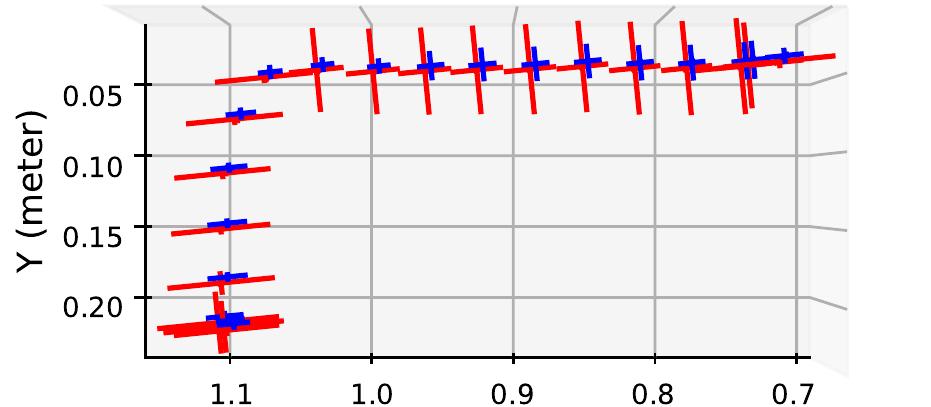}} \\
    \subfloat[With joint torque cost]{\includegraphics[trim={85 80 85 70},clip,width=0.44\textwidth]{exp_polish_jt.pdf}}\\
    \subfloat[Time plot of damping (top) and inertia (bottom)]{\includegraphics[trim={0 0 0 10},clip,width=0.35\textwidth]{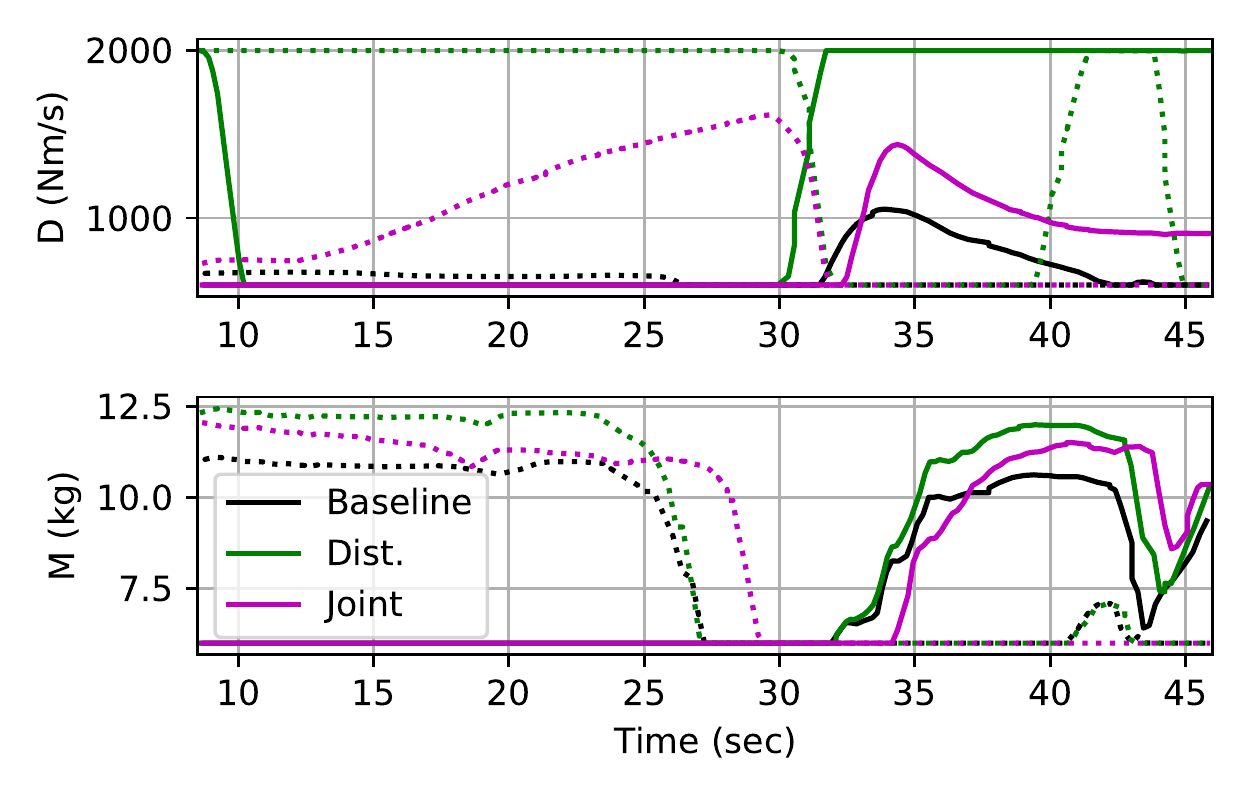}}\\
    \caption{Impedance values in the polishing task, showing that (b) frequency disturbance increases the mass across from the path, and (c) the human joint torque cost reduces impedance when farther from human, where more joint torque would be required. In (d) impedance over time is shown, where dotted is $y$-axis, solid $x$. The transition at $31$ seconds corresponds to the corner in the trajectory.}
    \label{fig:exp_polish}
\end{figure}
\begin{figure}[t!]
    \centering
    \includegraphics[width=0.47\textwidth]{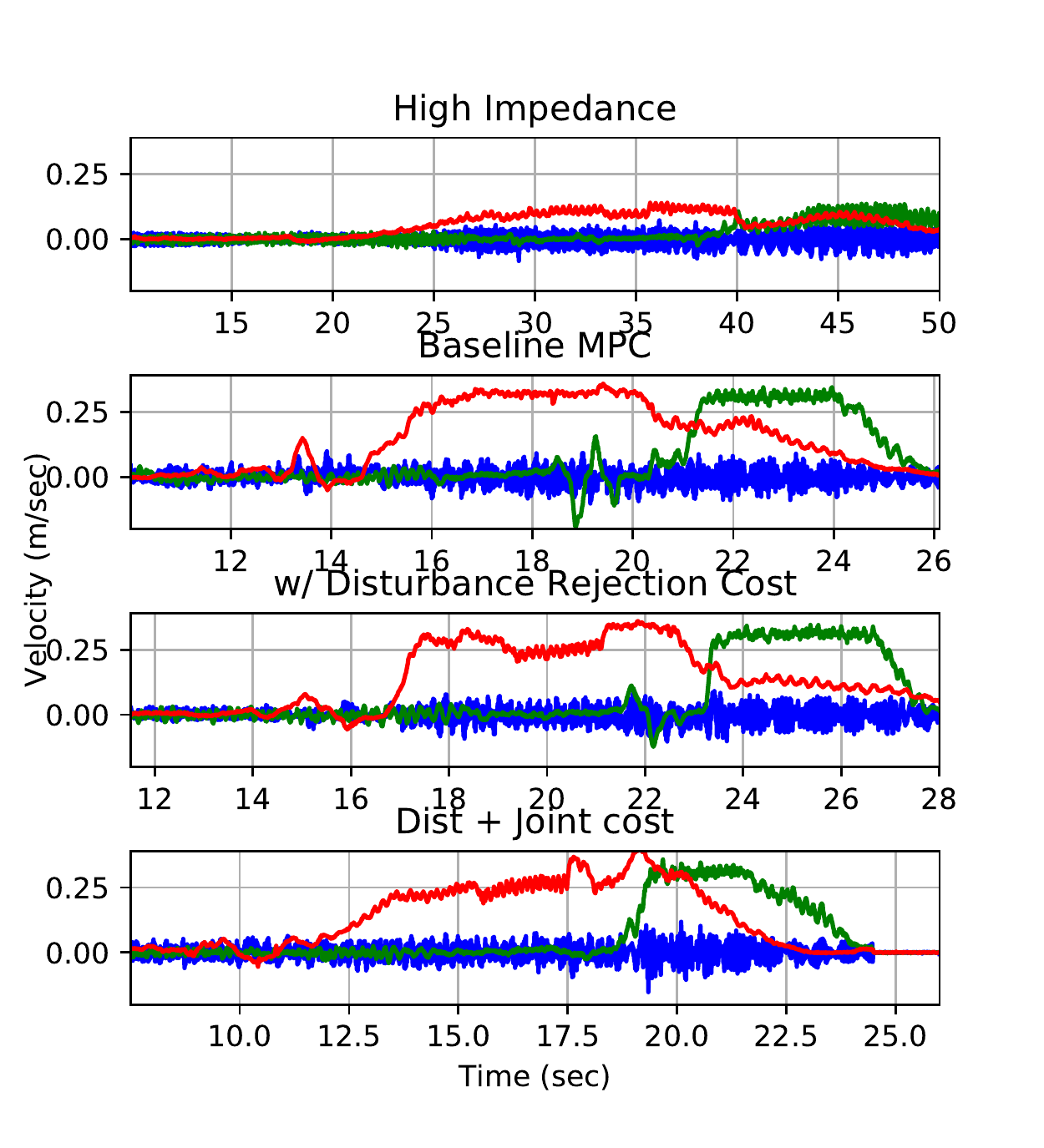} \\
    \caption{Velocity realized in polishing experiments, colors match coordinate system in Fig. \ref{fig:exp_polish_setup}. The standard cost function is compared with adding disturbance cost. It can be seen that the disturbance result decreases the high-frequency ($>$10 Hz) velocity of the robot.}
    \label{fig:exp_polish_velocity}
\end{figure}

\begin{table}[h]
\renewcommand{\arraystretch}{1.6} 
\begin{center}
\caption{RMS velocity below and above $15$ Hz, approximating the intentional motion and vibration, respectively. The high impedance and disturbance term suppress high-frequency vibration, but the high impedance results in a slow system.\label{tab:RMS_polish}}
\begin{tabular}{r | r r r r } 
MPC type & High imp. & Baseline & Dist. & Dist. + Jt. \\ \hline
High-freq RMS &  5.4e-6  & 7.3e-6 & 6.5e-6 & 7.6e-6  \\
Low-freq RMS & 0.8e-4 & 4.2e-4 & 4.2e-4 & 4.1e-4 \\
\end{tabular}
\end{center}
\end{table}

As seen in Fig. \ref{fig:exp_polish}, the different cost function terms affected the rendered impedance along the polishing path. When the frequency-domain disturbance was added, the directionality of the impedance was significantly affected, resulting in a higher impedance cross to the direction of motion, reducing oscillation cross from the direction of motion. This resulted in a reduction of high-frequency velocity, as seen in Fig. \ref{fig:exp_polish_velocity} and Table \ref{tab:RMS_polish}. Adding a penalty on human joint torques reduced the impedance when farther from the human, as an equivalent force at a greater distance requires more shoulder joint torques due to poor force manipulability. However, the human joint cost did not improve the vibration suppression as significantly as the disturbance cost as seen in Table \ref{tab:RMS_polish}. 

\subsection{Double Peg-in-Hole \label{sec:exp_pih}}

\begin{figure}[t!]
    \centering
    \subfloat[Experimental setup]{\includegraphics[width=0.47\textwidth]{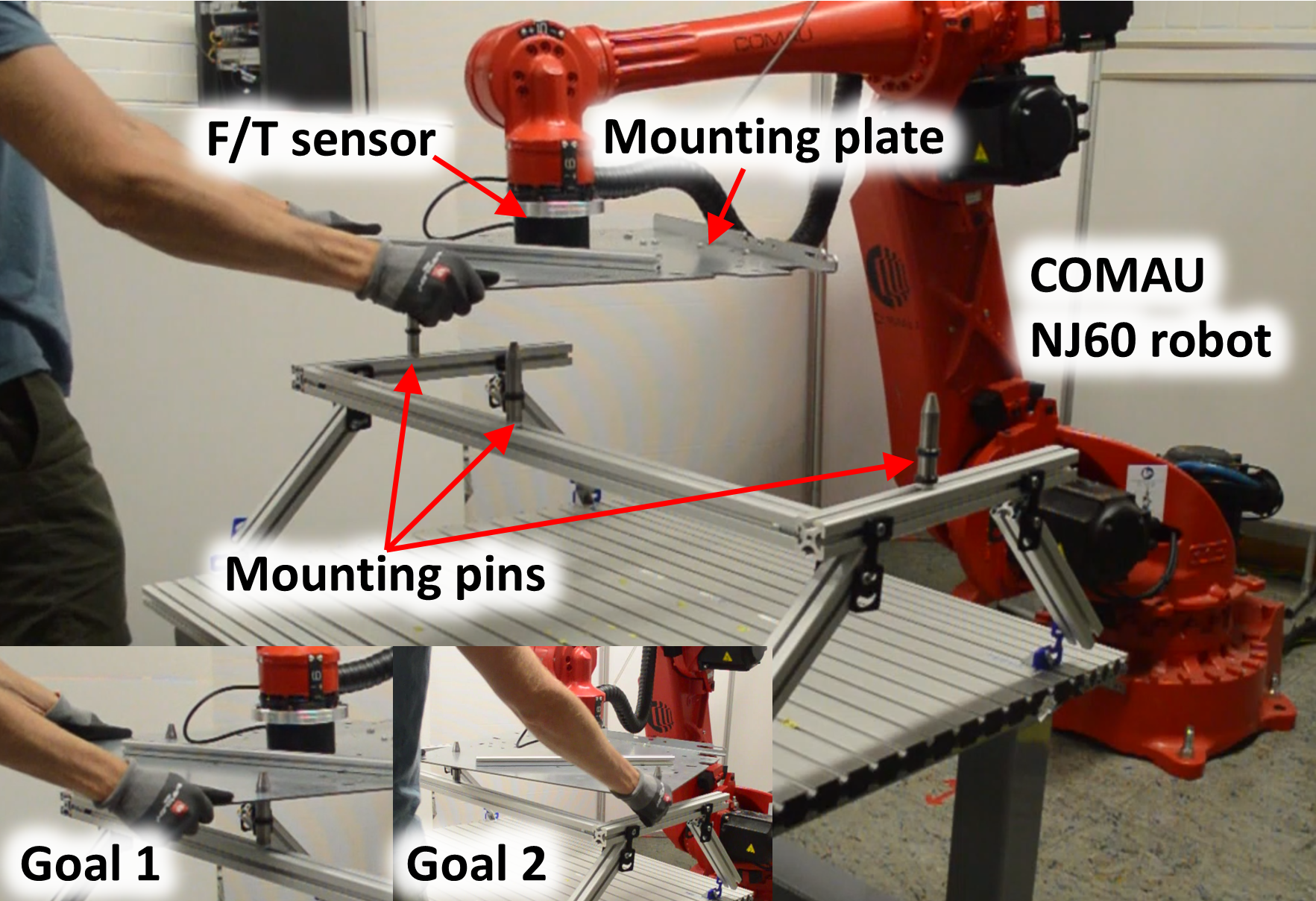}} \\
    \subfloat[Belief]{\includegraphics[trim={0 10 0 23},clip,width=0.47\textwidth]{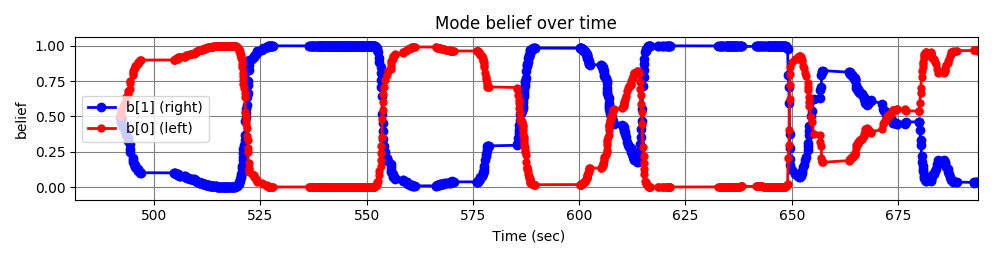}} \\
    \subfloat[Desired force trajectory]{\includegraphics[trim={0 10 0 23},clip,width=0.47\textwidth]{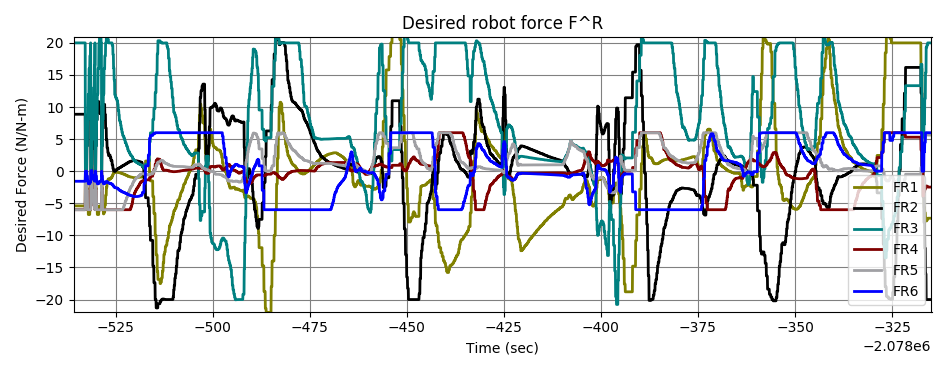}}
    \caption{Dual peg-in-hole task}
    \label{fig:exp_pih}
\end{figure}

This task involved multiple possible goals from the human, with the two goals indicated in Fig. \ref{fig:exp_pih}(a). While the human objective was not known in advance, the belief was updated online as seen in Fig. \ref{fig:exp_pih}(b). This update in the belief resulted in different force trajectories, as seen in Fig. \ref{fig:exp_pih}(c). This allowed the planned robot trajectory to respond online to changes in the human intent, providing 6-DOF assistance to the human.

\section{Conclusion}
This paper presented an MPC framework for impedance control in contact with humans and the environment. The use of Gaussian Processes to model external force allows explicit consideration of {\it variability in demonstrations} (rail assembly), {\it confidence in the skill} (higher covariance outside demonstrations), and uncertain {\it environment constraints} (Fig. \ref{fig:contact_var}). By using a GP model per human goal, a discrete goal can be estimated online to handle uncertainty in {\it human goal}. This approach can also consider frequency-domain disturbance models for {\it external perturbations}. By using stochastic trajectories, the impedance parameters can be directly optimized. 

While flexible, the approach has some limitations, mostly computational. Many of these limits arise from contact. While the hinge function in GPs allows better modeling of stiff contact, the integration over these stiff contact models requires a small step size, even with implicit integrators. Without suitable integrator parameters, the contact may be lost in the future trajectory and constraints such as chance or well-damped lose effectiveness. These inequality constraints are effectively solved in an interior-point method, but care must be taken to keep the problem feasible, i.e., allowing fast changes in damping to react to fast changes in projected environment stiffness. 

Further limitations arise from the GP models, which cannot model certain types of uncertainty--e.g., variation in noise model over the state space. Optimization of the GP parameters needs strong regularization, especially to keep comparable between DOF or different GP models.

\clearpage

\bibliographystyle{IEEEtran}
\bibliography{lib, lib_luka}

\end{document}